\documentclass[letterpaper, 10 pt, conference]{ieeeconf}
\IEEEoverridecommandlockouts
\usepackage{cite}
\usepackage{amsmath,amssymb,amsfonts}
\usepackage{algorithmic}
\usepackage{graphicx}
\usepackage{textcomp}
\usepackage{xcolor}
\usepackage{algorithm}
\usepackage{array}
\usepackage[caption=false,font=normalsize,labelfont=sf,textfont=sf]{subfig}
\usepackage{stfloats}
\usepackage{url}
\usepackage{verbatim}
\usepackage{times}
\usepackage{epsfig}
\usepackage{float}
\usepackage{wrapfig}
\usepackage{bm,xspace}
\usepackage{comment}
\usepackage{multirow}
\usepackage{balance}
\usepackage{booktabs}
\usepackage{etoolbox,siunitx}
\usepackage{calc}
\usepackage{pifont,hologo}
\usepackage{nicefrac}
\usepackage{makecell}
\usepackage{svg}
\usepackage{enumerate}

\makeatletter
\let\NAT@parse\undefined
\makeatother
\usepackage[colorlinks=black,citecolor=green]{hyperref}

\usepackage{adjustbox}
\def\BibTeX{{\rm B\kern-.05em{\sc i\kern-.025em b}\kern-.08em
    T\kern-.1667em\lower.7ex\hbox{E}\kern-.125emX}}

\begin{document}

\title{P-MapNet: Far-seeing Map Generator Enhanced by both SDMap and HDMap Priors}

\author{Zhou Jiang$^{1,2*}$, Zhenxin Zhu$^{2,3*}$, Pengfei Li$^{2,4}$, Huan-ang Gao$^{2,4}$, Tianyuan Yuan$^{4}$, Yongliang Shi$^{2}$,\\ Hang Zhao$^{4}$, Hao Zhao$^{2,\dag}$ % <-this % stops a space
\thanks{*Equal contribution, \dag  Corresponding author}% <-this % stops a space
\thanks{$^{1}$Beijing Institute of Technology, China, jzian@bit.edu.cn.}%
\thanks{$^{2}$Institute for AI Industry Research (AIR), Tsinghua University, China,
    \ zhaohao@air.tsinghua.edu.cn.}%
\thanks{$^{3}$Beihang University, China,
       zhuzhenxin@buaa.edu.cn.}%
\thanks{$^{4}$Tsinghua University, China.
      }%       
}

\makeatletter
\let\NAT@parse\undefined
\makeatother
\makeatletter
\g@addto@macro\@maketitle

\makeatother
\maketitle

\begin{abstract}

Autonomous vehicles are gradually entering city roads today, with the help of high-definition maps (HDMaps). However, the reliance on HDMaps prevents autonomous vehicles from stepping into regions without this expensive digital infrastructure. This fact drives many researchers to study online HDMap generation algorithms, but the performance of these algorithms at far regions is still unsatisfying. 
We present P-MapNet, in which the letter P highlights the fact that we focus on incorporating map priors to improve model performance. Specifically, we exploit priors in both SDMap and HDMap. On one hand, we extract weakly aligned SDMap from OpenStreetMap, and encode it as an additional conditioning branch. Despite the misalignment challenge, our attention-based architecture adaptively attends to relevant SDMap skeletons and significantly improves performance. On the other hand, we exploit a masked autoencoder to capture the prior distribution of HDMap, which can serve as a refinement module to mitigate occlusions and artifacts. 
We benchmark on the nuScenes and Argoverse2 datasets. Through comprehensive experiments, we show that: (1) our SDMap prior can improve online map generation performance, using both rasterized (by up to $+18.73$ $\rm mIoU$) and vectorized (by up to $+8.50$ $\rm mAP$) output representations. (2) our HDMap prior can improve map perceptual metrics by up to $6.34\%$. (3) P-MapNet can be switched into different inference modes that covers different regions of the accuracy-efficiency trade-off landscape. (4) P-MapNet is a  far-seeing solution that brings larger improvements on longer ranges. Codes and models are publicly available at 
\href{https://jike5.github.io/P-MapNet/}{https://jike5.github.io/P-MapNet/}.

\end{abstract}

%\begin{IEEEkeywords}
%NeRF, localization, low-pass filter.
%\end{IEEEkeywords}

\section{Introduction}
\begin{figure}[!t]
% \centering
\includegraphics[width=0.48\textwidth]{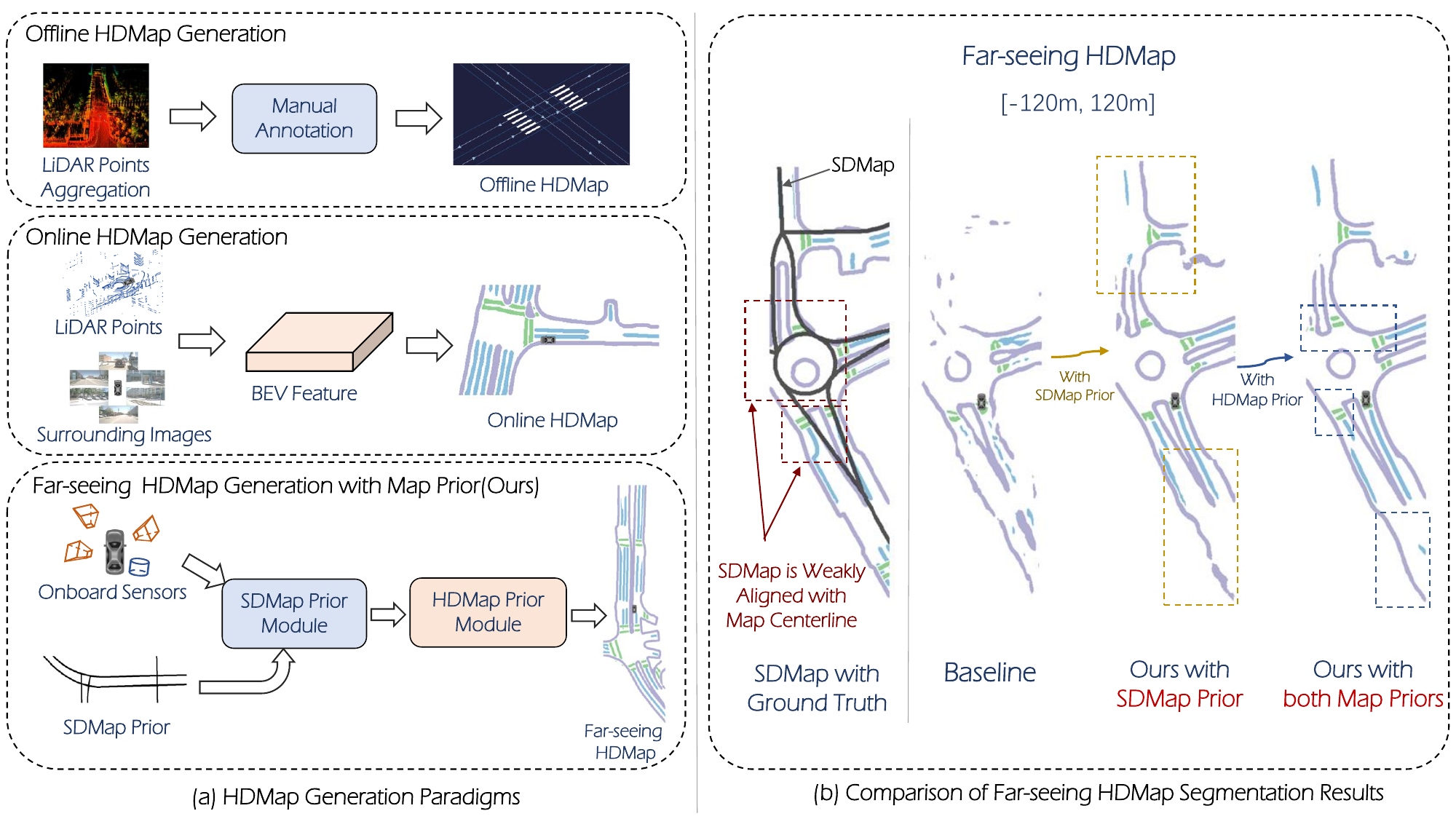}\caption{Left: Since offline HDMap generation is cumbersome and expensive, people are pursuing online HDMap generation algorithms and our P-MapNet is an online HDMap generator enhanced by both SDMap and HDMap priors. Right: Despite the misalignment between SDMaps and HDMaps, our P-MapNet can significantly improve map generation performance, especially on the far side.}
\label{teaser}
\vspace{-2mm}
\end{figure}

While we still don't know the ultimate answer to fully autonomous vehicles that can run smoothly in each and every corner of the earth, the community does have seen some impressive milestones, e.g., robotaxis are under steady operation in some big cities now. Yet current autonomous driving stacks heavily depend on an expensive digital infrastructure: HDMaps. With the availability of HDMaps, local maneuvers are reduced to lane following and lane changing coupled with dynamic obstacle avoidance, significantly narrowing down the space of decision making. But the generation of HDMaps, which is shown in the left-top panel of Fig.~\ref{teaser}, is very cumbersome and expensive. And what's worse, HDMaps cannot be generated for good and all, because they must be updated every three months on average. It is widely recognized that reducing reliance on HDMaps is critical.

Thus, several recent methods \cite{li2022hdmapnet,liu2022vectormapnet} generate HDMaps using multi-modal online sensory inputs like LiDAR point clouds and panoramic multi-view RGB images, and a conceptual illustration of this paradigm is given in the left-middle panel of Fig.~\ref{teaser}. Despite promising results achieved by these methods, long range distance online HDMap generators still report limited quantitative metrics and this study focuses on promoting their performance using priors. Specifically, two sources of priors are exploited: SDMap and HDMap, as demonstrated in the left-bottom panel of Fig.~\ref{teaser}.

\textbf{SDMap Prior.} Before the industry turns to build the digital infrastructure of HDMaps on a large scale, Standard Definition Maps (SDMaps) have been used for years and have largely promoted the convenience of our daily lives. Commercial SDMap applications provided by Google or Baidu help people navigate big cities with complex road networks, telling us to make turns at crossings or merge into main roads. SDMaps are not readily useful for autonomous cars because they only provide center-line skeletons (noted as SDMap Prior in the left-bottom panel of Fig.~\ref{teaser}). So we aim to exploit SDMap priors to build better online HDMap generation algorithms, which can be intuitively interpreted as \emph{drawing} HDmaps around the skeleton of SDMaps. However, this intuitive idea faces a primary challenge: misalignment. Per implementation, we extract SDMaps from OpenStreetMap using GPS signals but unfortunately they are, at best, weakly aligned with the ground truth HDMap in a certain scenario. An illustration is given in the right panel of Fig.~\ref{teaser}, noted as SDMap with Ground Truth. To this end, we leverage an attention based network architecture that adaptively attends to relevant SDMap features and successfully improve the performance by large margins in various settings (see Table.~\ref{tab:mIOU}). 

\textbf{HDMap Prior.} Although useful, SDMap priors cannot fully capture the distribution of HDMap output space. As noted by Ours with SDMap Prior in the right panel of Fig.~\ref{teaser}, HDMap generation results are broken and unnecessarily curved. This is credited to the fact that our architecture is, like prior methods, designed in a BEV dense prediction manner and the structured output space of BEV HDMap cannot be guaranteed. As such, HDMap prior comes to the stage as a solution and the intuition is that if the algorithm models the structured output space of HDMaps explicitly, it can naturally correct these unnatural artifacts (i.e., broken and unnecessarily curved results mentioned above). On the implementation side, we train a masked autoencoder (MAE) on a large set of HDMaps to capture the HDMap prior and use   it as a refinement module. As noted by ours with both Map Priors in the right panel of Fig.~\ref{teaser}, our MAE successfully corrects aforementioned issues.

\textbf{P-MapNet as a far-seeing solution.} A closer look at the positive margins brought by incorporating priors reveals that P-MapNet is a far-seeing solution. As shown in the right panel of Fig.~\ref{teaser}, after incorporating the SDMap prior, missing map elements far from the ego vehicle (denoted by the car icon) are successfully extracted. This is understandable as the road center-line skeletons on the far side are already known in the SDMap input. Meanwhile, the HDMap prior brings improvements in two kinds of regions: crossings with highly structured repetitive patterns and lanes on the far side. This is credited to the fact that our masked autoencoder can incorporate the priors about how typical HDMaps look like, e.g., lanes should be connected and largely straight and crossings are drawn in a repetitive manner. As Table.~\ref{tab:mIOU} demonstrates, positive margins steadily grow along with the sensing range. We believe P-MapNet, as a far-seeing solution, is potentially helpful in deriving more intelligent decisions that are informed of maps on the far side.

In summary, our contributions are three-fold: (1) We incorporate SDMap priors into online map generators by attending to weakly aligned SDMap features and achieve significant performance improvements; (2) We also incorporate HDMap priors using a masked autoencoder as a refinement module, correcting artefacts that deviate the structured output space of HDMaps; (3) We achieve state-of-the-art results of \textit{far-seeing} HDMap generation on public benchmarks and present in-depth ablative analyses revealing the mechanism. 
%For example, P-MapNet is a far-seeing solution.

\begin{figure*}[!t]
\centering
\vspace{3mm}
\includegraphics[width=0.8\textwidth]{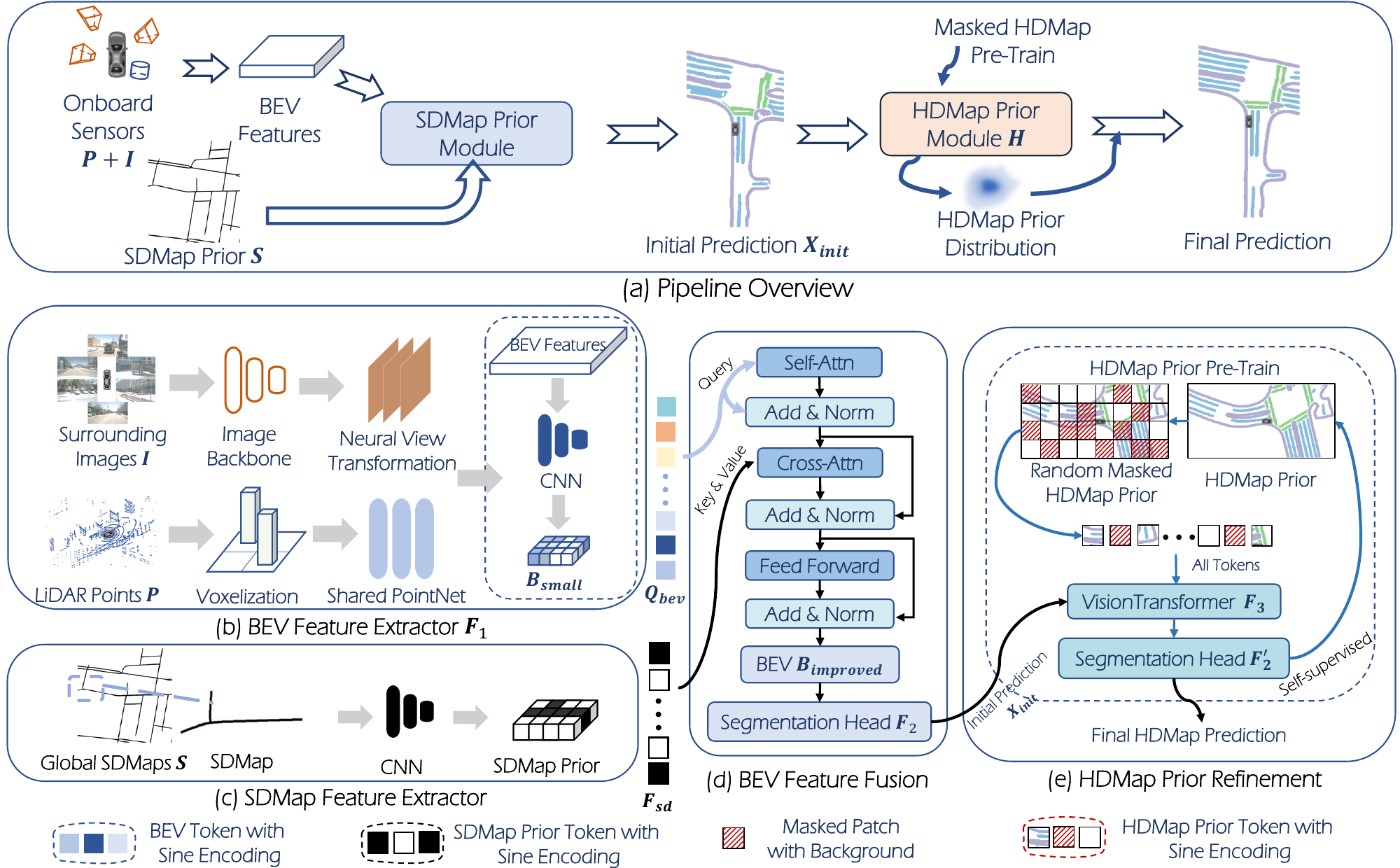}
\caption{\textbf{P-MapNet overview.} P-MapNet is designed to accept either surrounding cameras or multi-modal inputs. It processes these inputs to extract sensors features and SDMap priors features, both represented in the Bird's Eye View (BEV) space. These features are then fused using an attention mechanism and subsequently refined by the HDMap prior module to produce results that closely align with real-world map data.}
\label{main}
\vspace{-4mm}
\end{figure*}

\section{RELATED WORK}

\subsection{Online HD Map generation}
Online map generators are important for autonomous driving \cite{hu2023planning}\cite{wang2024openlane}\cite{zheng2023steps}\cite{wu2023mars}\cite{jin2023adapt}, which is similar in spirit to room layout estimation \cite{zhao2017physics}\cite{chen2022pqtransformer}\cite{gao2023semiOmni-PQ} for indoor scenes.
Traditionally, HD maps are manually annotated offline, combining point cloud maps via SLAM algorithms\cite{Bao2022HighDefinitionMG, houston2021one} for high accuracy but at a high cost and without real-time updates.
In contrast, recent efforts have focused on utilizing onboard sensors for the efficient and cost-effective generation of online HD maps\cite{LSS, saha2022translating, li2022hdmapnet, dong2022superfusion,liu2022vectormapnet, liao2023lane}. HDMapNet\cite{li2022hdmapnet} employs pixel-wise labelling and heuristic post-processing, using Average Precision (AP) and Intersection over Union (IoU) as metrics.
More recent approaches\cite{liao2022maptr, liu2022vectormapnet, ding2023pivotnet, yuan2023streammapnet, mapvr}, have adopted end-to-end vectorized HD map generation techniques, leveraging Transformer architectures\cite{vaswani2017attention}. 
However, these methods rely solely on onboard sensors and have limitations in handling challenging environmental conditions like occlusions or adverse weather.

\subsection{Long-range Map Perception}
To enhance the practicality of HD maps for downstream tasks, some studies aim to extend their coverage to longer perception ranges. SuperFusion\cite{dong2022superfusion} combines LiDAR point clouds and camera images for depth-aware BEV transformation, yielding forward-view HD map predictions up to $90~m$.
NeuralMapPrior\cite{xiong2023neuralmap} maintains and updates a global neural map prior, enhancing online observations to generate higher-quality, extended-range HD map predictions.
\cite{gao2023complementing} proposes using satellite maps to aid online map generation. Features from onboard sensors and satellite images are aggregated through a hierarchical fusion module to obtain the final BEV features.
MV-Map\cite{xie2023mvmap} specializes in offline, long-range HD map generation. It aggregates all relevant frames during traversal and optimizes neural radiance fields for improved BEV feature generation. 

\section{Formulation}

Given the LiDAR point cloud $\mathcal{P}$ and panoramic images $\left \{ \mathcal{I}_i \mid i = 1,2,\cdots N \right \}$, where $N$ is typically six for a panoramic rig, a common online HDMap generation task (e.g., HDMapNet \cite{li2022hdmapnet}) can be formulated as:
\begin{equation}
   \mathcal{M} = \mathcal{F}_2(\mathcal{F}_1(\mathcal{P},\mathcal{I})), \\
\end{equation}
where $\mathcal{F}_1$ represents the feature extractor that takes multi-modal inputs and generates BEV features, while $\mathcal{F}_2$ is a segmentation head that predicts a semantic category label for each grid in the BEV. And $\mathcal{M}$ is the HDMap prediction.

However, this common formualtion fails to leverage the rich priors in SDMap and HDMap. So we formulate a new task to incorporate these priors to produce a more accurate and \emph{far-seeing} HDMap, thereby effectively addressing issues pertaining to occlusion as well as super long range sensing:

%\textbf{New Task:}
%Given $\mathcal{P}$ and $\mathcal{I}$, the new task is to generate the \textit{far-seeing} HDMap, which is essential for the downstream long-term planning task.  

\begin{equation}
       \mathcal{M}^{'} = \mathcal{H}(\mathcal{F}_2(\mathcal{F}_1(\mathcal{P}, \mathcal{I}, \mathcal{S})))),
\end{equation}

Here $\mathcal{S}$ is the SDMap prior that comes in the form of road center-line skeletons. $\mathcal{H}$ represents the refinement module which is a pre-trained model capturing the distribution characteristics of HDMap. Similarly, $\mathcal{M}^{'}$ is the \textit{far-seeing} HDMap prediction over 100 meters on the front/back side. 

\textbf{Output Format.} There are two typical output formats for online HDMap generation: rasterized and vectorized. 
In this study, we focus on the rasterized representation (e.g. HDMapNet \cite{li2022hdmapnet}), as it is more suitable for designing our two prior modules (than vectorized counterpart). Specifically, how to effective encode vectorized representation for input/output is not as natural as rasterized representation.
% Here we primarily focus on rasterized representation because it is formulated in a dense prediction manner, thereby does not suffer from the label ambiguity issue. The vectorized representation, on the other hand, grapples with the challenge of defining an unambiguous vectorized representation. Having that said, we provide additional experiments, detailed in the supplementary materials, to demonstrate that our SDMap prior is also applicable to vectorized representation.
%For rasterized representation, we formualte the problem as such. 

\textbf{$\mathcal{S}$-only setting.} Shown in Fig. \ref{main}(a), we incorporate SDMap prior by encoding center-line skeletons as an additional input branch. In this $\mathcal{S}$-only setting, the formulation is: %Specifically, the SDMap Prior Module, designed as an additional input conditioning branch, 
 %incorporates SDMap features to enhance HDMap generation, . It can be formulated as:
 \begin{equation}
 \label{func: sd}
        \mathcal{M}_{S} = \mathcal{F}_2(\mathcal{F}_1(\mathcal{P}, \mathcal{I}, \mathcal{S})),\\
\end{equation}
 where the procedure of encoding $\mathcal{S}$ is illustrated in Fig. \ref{main}(c).
 
\textbf{$\mathcal{S}$+$\mathcal{H}$ setting.} While the SDMap prior $\mathcal{S}$ is naturally incorporated as an additional input, leveraging the HDMap prior is challenging. Our innovative proposal is to incorporate HDMap prior using a masked auto-encoder (MAE) as a refinement module. The core idea is to use MAE for the reconstruction of HDMaps so that this MAE intrinsically capture the distribution of HDMap prior. However, this is not trivial as the vanilla MAE cannot achieve this goal. 

\textbf{Vanilla MAE.} A vanilla MAE \cite{he2022mae} would treat HDMaps as images and trains under an MSE loss for image reconstruction. The problem is that this MAE would predict images thus cannot be used as a refinement module, as our HDMap generator actually needs a segmentation head at last.

\textbf{Our MAE variant.} Our MAE variant takes rasterized HDMap (which is in nature images) as input, but predicts the semantic label of each grid using a segmentation head. This is still an auto-encoding process since the module reconstructs the HDMap of interest. However, this MAE's input and output comes in different formats: images and segmentation masks. This readily allows refinement when attached after the $\mathcal{M}_{S}$ output mentioned above. %is designed as a pre-trained post-processing refinement step to obtain a more realism \textit{far-seeing} HDMap. We utilize a novel masked autoencoder to capture the distribution of HDMap to make the initial HDMap prediction be more accurate and realism, as depicted in Fig. \ref{main}(a).

\textbf{Formal notation.} our HDMap refinement module has two training steps. The first step is to pre-train the HDMap Prior Module on a large set of HDMaps, as shown in Fig. \ref{main}(e).
\begin{equation}
\begin{aligned}
\label{func: self}
 \mathcal{M}_{\rm self} & =  \mathcal{H}(\mathcal{M}_{\rm masked}) \\
 & \triangleq \mathcal{F}_2^{'}(\mathcal{F}_3(\mathcal{M}_{\rm masked})),
 \end{aligned}
\end{equation}
Here the HDMap Prior Module $\mathcal{H}(\cdot)$ is specifically defined as $\mathcal{F}^{'}_2(\mathcal{F}_3(\cdot))$, of which $\mathcal{F}_3$ denotes a ViT model used in typical MAEs. But $\mathcal{F}^{'}_2$ denotes another segmentation head, as mentioned above. This  $\mathcal{F}^{'}_2$ makes our MAE a variant ready for refinement. $\mathcal{M}_{\rm masked}$ is the randomly masked HDMap from the training dataset and $\mathcal{M}_{\rm self}$ is the unmasked version.

\textbf{Fine-tuning.} The second step is end-to-end fine-tuning:
\begin{equation}
\begin{aligned}
\label{func: refine}
       \mathcal{M}^{'} &=\mathcal{H}(\mathcal{M}_{S})\\
        &\triangleq \mathcal{F}^{'}_2(\mathcal{F}_3(\mathcal{M}_{S})),
\end{aligned}
\end{equation}
where $\mathcal{M}_{S}$ is the initial prediction from SDMap Prior Module, as depicted in Fig. \ref{main}(a) and Equation.3.

As such, the formal $\mathcal{S}$+$\mathcal{H}$ setting (by integrating Equation.3 and Equation.5) is shown as:

\begin{equation}
 \label{func: sd}
        \mathcal{M}^{'} = \mathcal{F}^{'}_2(\mathcal{F}_3(\mathcal{F}_2(\mathcal{F}_1(\mathcal{P}, \mathcal{I}, \mathcal{S}))))
\end{equation}
% \begin{equation}
%        \mathcal{M}_{H} = \mathcal{F}^{'}_2(\mathcal{F}_3(\mathcal{F}_2(\mathcal{F}_1(\mathcal{P}, \mathcal{I}, \mathcal{S}))))
% \end{equation}

% \textbf{Formulation:} In the formulation, $\mathcal{F}_1$ represents the feature extractor, $\mathcal{F}_2$ and $\mathcal{F}^{'}_2$ are two segmentation head with the same network architecture, $\mathcal{F}_3$ and $\mathcal{F}^{'}_2$ constitute the MAE module, where $\mathcal{F}_3$ is a pretrained ViT model.

% Baseline method extract BEV features from LiDAR point clouds $\mathcal{P}$ and surrounding images $\mathcal{I}$ and then process through a segmentation head to generate the rasterized HDMap.

% SDMap Prior Module incorporates SDMaps features as an additional inputs to enhance the HDMap generation.

% HDMap Prior Module first utilizes self-supervised learning to pretrain the masked autoencoder via a ViT\cite{dosovitskiy2020ViT} model and a segmentation head $\mathcal{F}_2^{'}$, which is formulated as $\mathcal{M}_{self} = \mathcal{F}_2^{'}(\mathcal{F}_3(\mathcal{M}_{masked}))$, where $\mathcal{M}_{masked}$ is the masked HDMap and $\mathcal{M}_{self}$ is the completed map to make the self-supervised learning to capture the HDMap distribution.

% Notably, our two modules strike a good balance between efficiency and accuracy, the detailed break-down run time is demonstrated in Fig. \ref{Fig:time}.

\section{METHOD}

\subsection{SDMap Prior Module} \label{sdmap}
Now we elaborate on the implementation of our SDMap prior module, and first we recap the motivation again: Given the intrinsic challenges associated with onboard sensing, such as distant road invisibility and adverse weather conditions, incorporating SDMap prior becomes a promising technique, as SDMap provides a stable and consistent outlining of the environment (agnostic of those challenges).

\textbf{SDMap Generation.}
We first introduce our approach to generate SDMap priors by leveraging OpenStreetMap (OSM)\cite{haklay2008openstreetmap} data. 
We specifically employ the nuScenes \cite{caesar2020nuscenes} and Argoverse2 \cite{Argoverse2} datasets for our research, as these datasets hold a prominent position within the autonomous driving domain. These datasets are richly equipped with sensors but do not include the corresponding SDMap information for the captured regions. To address this limitation, we leverage OpenStreetMap to obtain the relevant SDMap data for these regions.
Specifically, we first obtain the localized SDMap data of the corresponding area from the OSM\footnote{https://www.openstreetmap.org/} based on the on-board GPS information. Then these SDMap data are transformed to the ego vehicle's coordinate system.  Although we obtain the SDMap priors, the problem of misalignment due to the low accuracy of the OSM and the bias of the GPS will pose a challenge to the fusion of SDMap priors.

\textbf{Incorporating SDMap Prior.} 
After extraction and rasterization, the rasterized SDMap prior inevitably faces spatial misalignment, where the SDMap prior doesn't align precisely with the current operational location, often resulting from inaccurate GPS signals or rapid vehicle movement.
This misalignment renders the straightforward method of directly concatenating BEV features with SDMap features in the feature dimension ineffective, as detailed in Tab.\ref{tab:fusion_mode}.
To tackle this challenge, we adopt a multi-head cross-attention module. This allows the network to utilize cross-attention to determine the most suitably aligned location, thereby effectively enhancing the BEV feature with the SDMap prior.

\textbf{BEV Query.} As illustrated in Fig.~\ref{main}(b), we first utilize a convolutional network to downsample the BEV features. This not only averts excessive memory consumption on low-level feature maps but also partially alleviates the misalignment between the image BEV features and the LiDAR BEV features. The downsampled BEV features are represented as \(\mathcal{B}_{\text{small}} \in \mathbb{R}^{\frac{H}{d} \times \frac{W}{d} \times C}\), where \(d\) is the downsampling factor. These features, combined with sine positional embedding and squeezed into 1D, resulting in the BEV queries \(\mathcal{Q}_{\text{bev}}\). 

\textbf{SDMaps Prior attended.} The associated (though misaligned) SDMap undergoes processing via a convolutional network in conjunction with sine positional embedding, producing the SDMap prior tokens \(\mathcal{F}_{\text{sd}}\), as shown in Fig.~\ref{main}(c). Subsequently, the multi-head cross-attention is employed to enhance the BEV queries by integrating the information from SDMap priors. The formal representation is,
\begin{equation}
\begin{gathered}
    \mathcal{Q}^{\prime} = \operatorname{Concat}\left(\operatorname{CA}_{1}(\mathcal{Q}_{\text{bev}},\mathcal{F}_{\text{sd}}), \ldots, \operatorname{CA}_{m}(\mathcal{Q}_{\text{bev}},\mathcal{F}_{\text{sd}})\right),  \\
    \mathcal{B}_{\text{improved}} = \operatorname{layernorm}\left(\mathcal{Q}_{\text{bev}} + \operatorname{Dropout}(\operatorname{Proj}(\mathcal{Q}^{\prime})\right)),
\end{gathered}
\end{equation}
where the $\operatorname{CA}_i$ is the $i$-th single head cross-attention, $m$ is the number of head, key and value embeddings, $\operatorname{Proj}$ is a projection layer and $ \mathcal{B}_{\text{improved}} $ represents the resized BEV feature derived from the multi-head cross-attention that incorporates the SDMap prior.
Subsequently, the improved BEV features pass through a segmentation head to get the initial HDMap element prediction, denoted as \({X}_{\text{init}} \in \mathbb{R}^{H \times W \times (N_c+1)} \). Here, the $(N_c+1)$ channels denote the total number of map element classes, including an additional background class.

\subsection{HDMap Prior Module}
\label{hdmap}

Then we describe the HDMap prior module, which $\mathcal{H}$ is computationally heavy (see Fig.\ref{Fig:time}) which is indeed optional. Our goal is to acquire a more precise and realistic \textit{far-seeing} HDMap, particularly in challenging scenarios such as inclement weather, areas of occlusion, and regions of invisibility. To enhance the continuity and realism of HDMap generation in these scenarios, closely approximating the distribution of HDMap, we employ an adapted pre-trained MAE module to capture the distribution. Training the HDMap Prior Module has two training steps: the first step involves training the MAE module using self-supervised learning to capture the HDMap distribution and the second step is fine-tuning by loading weights in the first step and using the 
initial HDMap prediction ${X}_{\text{init}}$ as input, as shown in Function. \ref{func: refine}.

\begin{figure}[!h]
    \centering
    \includegraphics[width=0.75\linewidth]{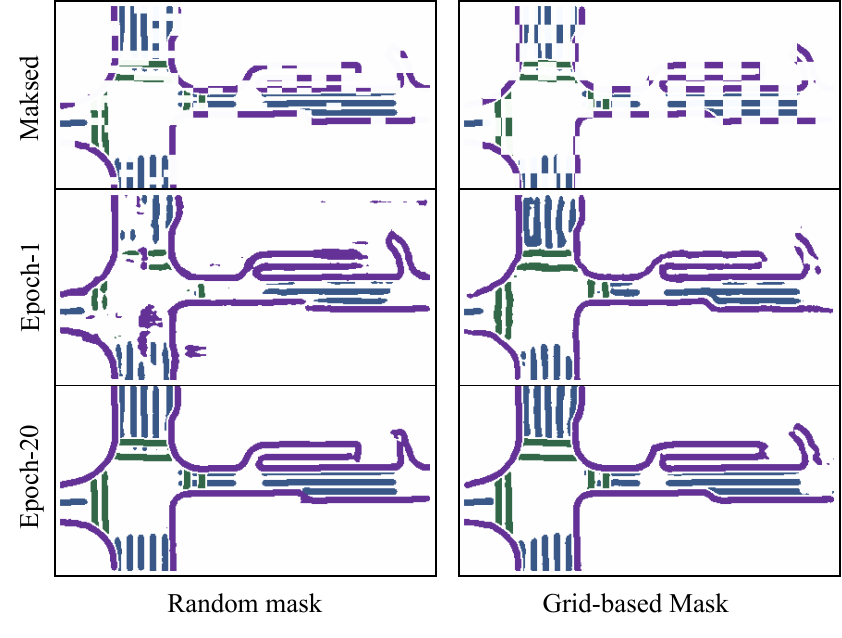}
    \vspace{-3mm}
    \caption{\textbf{Different mask strategies. }``Masked" refers to the pre-training inputs after applying various masking strategies, and ``Epoch-1" and ``Epoch-20" denote the reconstruction results at the first and twentieth epochs of the pre-training process, respectively.}
    \label{fig:mask}
\end{figure}

\textbf{Pre-trained MAE module.}
We utilize self-supervised learning to pre-train the masked autoencoder to capture the data distribution of HDMap. As shown in Fig. \ref{main}(e), this module consists of a Vision Transformer model\cite{dosovitskiy2020ViT} and a fully convolutional segmentation head. As illustrated in Function. \ref{func: self}, 
we mask the HDMap ground truth in the \textit{training} set of the dataset, and then encode this masked HDMap using the ViT model. Subsequently, given that our reconstruction target is indeed a semantic mask (although treated as images for input), we employ the segmentation head to revert the masked HDMap back to its original HDMap ground truth. This process is self-supervised using pixel-wise cross-entropy loss between the HDMap ground truth and the masked HDMap.
Specifically, we tried two different mask strategy to pre-train the module, namely grid mask and random mask, as shown in Fig. \ref{fig:mask}. In the random mask strategy, we randomly select both the mask patch size and the mask ratio from a set of candidates to mitigate the over-fitting issue during pre-training.

% This self-supervised approach enables the MAE module to learn and capture the HDMap distribution.

% Notably, different from a conventional MAE module, we leverage a segmentation head to complete the masked HDMap achieving the self-supervised leaning to capture the HDMap distribution. 

\textbf{End-to-end fine-tuning.}
Next we apply the pre-trained MAE module on the initial HDMap prediction ${X}_{\rm init}$ as a refinement plug-in to improve the initially predicted HDMap, addressing potential issues such as broken or missing lanes in challenging scenarios. We then take the entire model through some lightweight fine-tuning for 10 epochs to better align the distribution of the initial prediction with that of the HDMap distribution, as illustrated in Function \ref{func: refine}.

% to m 

% \textbf{Inference.}
% The entire model with both SDMap prior and HDMap prior is formulated as Function. \ref{func: all}, which is the comination of Function. \ref{func: sd} and Function. \ref{func: refine}. 
% \begin{equation}
% \label{func: all}
%        \mathcal{M}^{'} = \mathcal{F}^{'}_2(\mathcal{F}_3(\mathcal{F}_2(\mathcal{F}_1(\mathcal{P}, \mathcal{I}, \mathcal{S}))))
% \end{equation}
% where the $\mathcal{M}^{'}$ is the final HDMap prediction and the $\mathcal{F}_2$ and $\mathcal{F}^{'}_2$ is two different fully convolutional segmentation head with the same network.

% Furthermore, to demonstrate the potential of scalable training, 
% %in order to verify the generalization of the pre-trained MAE module and to avoid the over-fitting issue
% we conduct the cross-data ablation study, which is pre-trained in the Argoverse v2 train set and then refine in the NuScenes dataset, as shown in Tab. \ref{crossdata}.  
% More ablation studies are detailed in the supplementary materials to demonstrate the efficiency of the HDMap Prior Module.
\begin{figure*}[!t]
\centering
\vspace{2mm}
\includegraphics[width=0.80\textwidth]{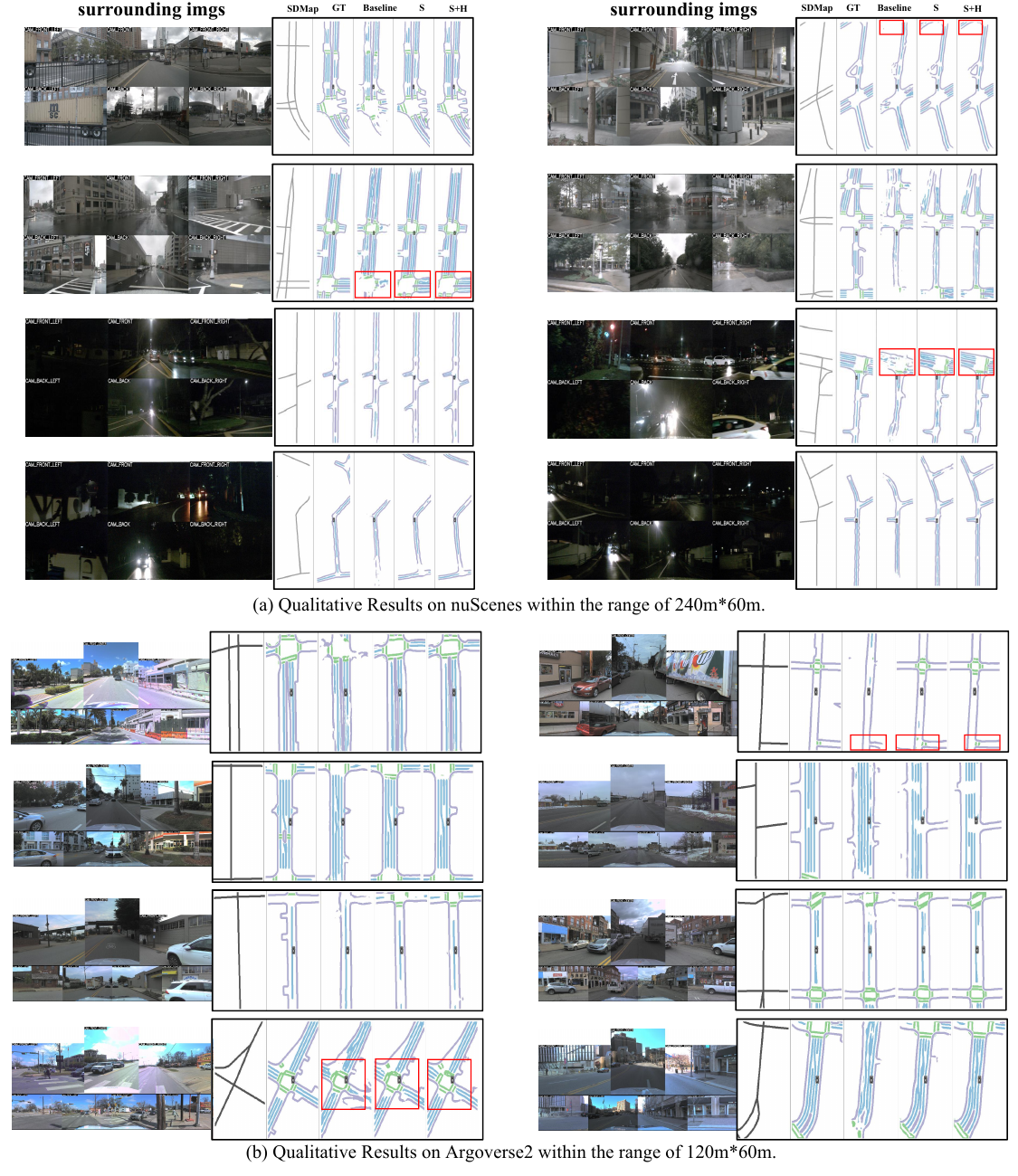}
\caption{\textbf{Qualitative results.} 
We conduct a comparative analysis within a range of 240m$\times$60m on nuScenes dataset and 120m$\times$60m on Argoverse2 dataset, utilizing C+L as input. In our notation, ``S" indicates that our method utilizes only the SDMap priors, while ``S+H" indicates the utilization of both. Our method consistently outperforms the baseline method under various weather conditions and in scenarios involving viewpoint occlusion.}
\label{Fig:qualitative_res}
\vspace{-2mm}
\end{figure*}

\begin{table*}[t]
  \vspace{3mm}
  \caption{\textbf{Quantitative results of IoU scores and AP scores.} Performance comparison of HDMapNet \cite{li2022hdmapnet} baseline and ours on the nuScenes val set \cite{caesar2020nuscenes}. ``S" indicates that our method utilizes only the SDMap priors, while ``S+H" indicates the utilization of the both priors. ``M" represents the Modality of our method and  ``Epoch" represents the number of refinement epochs.}
  \vspace{-6mm}
  \begin{center}
  \resizebox{1.0\textwidth}{!}{
    \begin{tabular}{c|c|cccc|cccc|cccc|c}
    \toprule
    Range & Method & S & S+H & M & Epoch & Div. & Ped. & Bound. & mIoU & Div. & Ped. & Bound. & mAP & FPS\\
    \midrule
    \multirow{5}{*}{$60\times 30\,m$} & HDMapNet &    &    & C & 30 & 40.5 & 19.7 & 40.5 & 33.57& 27.68 & 10.26 & 45.19 & 27.71 & 35.4\\

    \multirow{5}{*}{ } & P-MapNet & \checkmark &    & C & 30 & 44.1 & 22.6 & 43.8 &  36.83 \color{red}{(+3.26)} & 32.11 & 11.33 & 48.67 &  30.70 \color{red}{(+2.99)} & 30.2 \\
    \multirow{5}{*}{ } & P-MapNet & \checkmark &  \checkmark  & C & 10 & 44.3 & 23.3 & 43.8  & 37.13  \color{red}{(+3.56)}  & 26.08 & 17.66 & 48.43 &  30.72 \color{red}{(+3.01)}   & 12.2 \\
    \multirow{5}{*}{ } & HDMapNet &    &    & C+L & 30 & 45.9 & 30.5 & 56.8 & 44.40 & 29.46 & 13.89 & 54.07 & 32.47 & 21.4\\
    \multirow{5}{*}{ } & P-MapNet & \checkmark &    & C+L & 30 & 53.3 & 39.4 & 63.1 & 51.93 \color{red}{(+7.53)}  & 36.56 & 20.06 & 60.31 & 38.98 \color{red}{(+6.51)}  & 19.2 \\
    \multirow{5}{*}{ } & P-MapNet & \checkmark & \checkmark & C+L & 10 & \textbf{54.2} & \textbf{41.3} & \textbf{63.7} & \textbf{53.07 \color{red}{(+8.67)}}  &\textbf{37.81} & \textbf{24.96} & \textbf{60.90} & \textbf{41.22 \color{red}{(+8.75)}} & 9.6 \\
    \midrule
    \multirow{5}{*}{$120\times 60\,m$} & HDMapNet &    &    & C & 30 & 39.2 & 23.0 & 39.1 & 33.77& 14.40 & 8.98 & 34.99 & 19.46 & 34.2 \\

    \multirow{5}{*}{ } & P-MapNet & \checkmark &    & C & 30 &44.8  &30.6  &45.6  &40.33 \color{red}{(+6.56)} & 19.39 & 14.59 & 38.69 & 24.22 \color{red}{(+4.76)}&  28.7\\
    \multirow{5}{*}{ } & P-MapNet & \checkmark &  \checkmark  & C & 10 & 45.5 & 30.9 & 46.2 & 40.87 \color{red}{(+7.10)} &19.50 & 24.72 & 42.48 & 28.90 \textbf{\color{red}{(+9.44)}} & 12.1 \\
    \multirow{5}{*}{}  & HDMapNet &    &    & C+L & 30 & 53.6 & 37.8 & 57.1 & 49.50  & 21.11 & 18.90 & 47.31 & 29.11& 21.2\\
    \multirow{5}{*}{ } & P-MapNet & \checkmark &    & C+L &30 & 63.6 & 50.2 & 66.8 & 60.20 \color{red}{(+10.70)} & 28.30 & 25.67 & 52.51 & 35.49 \color{red}{(+6.38)} & 18.7\\
    \multirow{5}{*}{ } & P-MapNet & \checkmark & \checkmark & C+L & 10 & \textbf{65.3} & \textbf{52.0} & \textbf{68.0} & \textbf{61.77 \color{red}{(+12.27)}}& \textbf{30.63} & \textbf{28.42} & \textbf{53.27} & \textbf{37.44} 
 \color{red}{(+8.33)} & 9.6 \\
    \midrule
    \multirow{5}{*}{$240\times 60\,m$} & HDMapNet &    &    & C & 30 & 31.9 & 17.0 & 31.4 & 26.77 & 7.37 & 5.09 & 21.59 & 11.35 & 22.3 \\

    \multirow{5}{*}{} & P-MapNet & \checkmark &    & C & 30 &  46.3 & 35.7  & 44.6 &  42.20 \color{red}{(+15.43)} & 10.86 & 12.74 & 25.52 & 16.38 \color{red}{(+5.03)}& 19.2 \\ 
    \multirow{5}{*}{} & P-MapNet & \checkmark &  \checkmark  & C & 10 &  49.0 & 40.9  & 46.6 &  45.50 \textbf{\color{red}{(+18.73)}} & 14.51 & \textbf{25.63} & 28.11 & 22.75 \textbf{\color{red}{(+11.40)}}& 9.1 \\ 
    \multirow{5}{*}{} & HDMapNet &    &    & C+L & 30 & 40.0 & 26.8 & 42.6 & 36.47 & 11.29 & 11.40 & 29.05 & 17.25 &  13.1\\
    \multirow{5}{*}{} & P-MapNet & \checkmark &    & C+L & 30 & 52.0  & 41.0  & 53.6  & 48.87 \color{red}{(+12.40)} & 17.87  & 20.00  & \textbf{35.89} & 24.59 \color{red}{(+7.34)} & 10.9 \\ 
    \multirow{5}{*}{} & P-MapNet & \checkmark & \checkmark & C+L & 10 & \textbf{53.0} & \textbf{42.6} & \textbf{54.2} & \textbf{49.93} \color{red}{(+13.46)}& \textbf{21.47} & {24.14} & {34.23} & \textbf{26.61} \color{red}{(+9.36)} & 6.6\\
    
    \bottomrule
  \end{tabular}
  }
  \end{center}
  \label{tab:mIOU}
  \vspace{-6mm}
\end{table*}
\begin{table}[h]

  % \vspace{-4mm}
  \caption{\textbf{P-MapNet achieves state-of-the-art on NuScenes \textit{val} set.} The symbol ``\textdagger" denotes results reported in \cite{xie2023mvmap}, \cite{dong2022superfusion}, while ``NMP" represents the ``HDMapNet+NMP" configuration as described in \cite{xiong2023neuralmap}. For superlong-range perception, we compared with SuperFusion\cite{dong2022superfusion} and BEVFusion\cite{liu2022bevfusion}. ``C" and ``L" respectively refer to the surround-view cameras and LiDAR inputs.  Ours uses both SDMap and HDMap priors. }
  \begin{center}
  \resizebox{0.48\textwidth}{!}{
    \begin{tabular}{c|cc|ccc|c}
    \toprule
    Range & Method & Modality & Div. & Ped. & Bound. & mIoU\\
    \midrule
    \multirow{7}{*}{$60\times 30\,m$} & VPN\cite{pan2020vpn} \textdagger & C & 36.5 & 15.8 & 35.6 & 29.30 \\
    \multirow{7}{*}{ } & Lift-Splat-Shoot\cite{LSS} \textdagger & C & 38.3 & 14.9 & 39.3 & 30.83 \\
    \multirow{7}{*}{ } & HDMapNet\cite{li2022hdmapnet} & C & 40.5 & 19.7 & 40.5 & 33.57 \\
    \multirow{7}{*}{ } & NMP\cite{xiong2023neuralmap} \textdagger & C & 44.1 & 21.0 & \textbf{46.1} & 37.05 \\
    \multirow{7}{*}{ } & HDMapNet & C+L & 45.9 & 30.5 & 56.8 & 44.40 \\
    \multirow{7}{*}{ } & P-MapNet (Ours) & C & \textbf{44.3} & \textbf{23.3} & {43.8} & \textbf{37.13} \\
    \multirow{7}{*}{ } & P-MapNet (Ours) & C+L & \textbf{54.2} & \textbf{41.3} & \textbf{63.7} & \textbf{53.07} \\
    \midrule
    \multirow{3}{*}{$90 \times 30\,m$} & BEVFusion\cite{liu2022bevfusion} \textdagger & C+L & 33.9 & 18.8 & 38.8 & 30.50\\
    \multirow{3}{*}{ } & SuperFusion\cite{dong2022superfusion} \textdagger & C+L & 37.0 & 24.8 & 41.5 & 34.43 \\
    \multirow{3}{*}{ } & P-MapNet (Ours) & C+L & \textbf{44.73} & \textbf{31.03} & \textbf{45.5} & \textbf{40.64} \\
    \bottomrule
  \end{tabular}
  }
      
  \end{center}
  \label{tab:sota_method}
  \vspace{-8mm}
\end{table}
\section{EXPERIMENTS}

\subsection{Dataset and Metrics}
We evaluate P-MapNet on two popular datasets in autonomous driving research, nuScenes\cite{caesar2020nuscenes} and Argoverse2\cite{Argoverse2}. 
To demonstrate our method is a far-seeing solution, we set three distinct perception ranges along the direction of vehicle travel:
$60\times 30\,m$, $120\times 60\,m$, $240\times 60\,m$. 
% $[-30\mathrm{m}, 30\mathrm{m}]$, $[-60\mathrm{m}, 60\mathrm{m}]$, $[-120\mathrm{m}, 120\mathrm{m}]$
Additionally, we utilize different map resolutions, specifically 0.15m for the short range of $60\times 30\,m$ and 0.3m for the rest two longer ranges. We use intersection-over-union (IoU) as the metrics for segmentation results and incorporate a post-processing step to get the vectorized map and evaluate it using the average precision (AP). Following \cite{dong2022superfusion}, we set the threshold of IoU as 0.2 and threshold of CD as 0.5m, 1.0m, 1.5m. 
Furthermore, to evaluate the \textit{realism} of the HDMap prior refinement module output, we utilize a perceptual metric LPIPS\cite{Lpips}, which leverages deep learning techniques to more closely simulate human visual perception differences, providing a more precise and human vision-aligned image quality assessment than traditional pixel-level or simple structural comparisons. Implementation details can be found in the supplementary material.

% \subsection{Implementation Details}
% P-MapNet is trained with four NVIDIA GeForce RTX 3090 GPUs. We use the Adam\cite{kingma2014adam} optimizer and StepLR schedule for trainning with a learning rate of $5 \times 10^{-4}$.
%For fairness comparison, we adopt the EfficientNet-B0 \cite{tan2019efficientnet} pretrained on ImageNet\cite{russakovsky2015imagenet} as the perspective view image encoder and use a MLP to convert to the BEV features. 
%To encode the point clouds for the LiDAR BEV features, we utilize the PointPillars framework\cite{lang2019pointpillars}, operating at an output dimension of 128.
%During the pretraining phase for the HDMap prior, we trained for 20 epochs for each range. Subsequently, we combined the \textit{BEV Feature Fusion} with the \textit{HDMap Prior Refinement} module and conducted an additional 10 epochs of training to obtain the final HDMap predictions. 

\subsection{Results}

\textbf{Comparisons with State-of-the-arts.} 
We conducted a comparative analysis of our approach with current state-of-the-art (SOTA) approaches in both short-range ($60m \times 30m$) perception and long-range ($90m \times 30m$) with a resolution of 0.15m. As indicated in Tab.~\ref{tab:sota_method}, our method exhibits superior performance compared to both existing vision-only and multi-modal  (RGB+LiDAR) methods.
% Specifically, our method conducted experiments at 0.3m resolution for long-range perception, after which we upsample the predictions to 0.15m resolution and apply certain morphological operations to ensure a fair comparison.

\textbf{Far-seeing Experiments.} 
We performed a performance comparison with HDMapNet\cite{li2022hdmapnet} at various distances and using different sensor modalities, with the results summarized in Tab.~\ref{tab:mIOU} and Tab.~\ref{tab:av2_miou}. Our method achieves a remarkable 13.4\% improvement in mIOU at a range of $240m \times 60m$. It is noteworthy that the effectiveness of SD Map priors becomes more pronounced as the perception distance extends beyond or even surpasses the sensor detection range, thus validating the efficacy of SD Map priors.
Lastly, our utilization of HD Map priors contributes to additional performance improvements by refining the initial prediction results to be more realistic and eliminating results that are broken and unnecessarily
curved, as demonstrated in Fig.~\ref{Fig:qualitative_res}.

\begin{table}[h]
  \vspace{-1mm}
  \caption{\textbf{Quantitaive results of map segmentation on Argoverse2 \textit{val} set.} We conducted a comparison between the P-MapNet method and HDMapNet\cite{li2022hdmapnet}, using only surround-view cameras as input, demonstrating superior performance.}
  \begin{center}
  \resizebox{0.48\textwidth}{!}{
    \begin{tabular}{c|c|ccc|c}
    \toprule
    Range & Method  & Div. & Ped. & Bound. & mIoU\\
    \midrule
    \multirow{3}{*}{$60\times 30\,m$} & HDMapNet &  53.0 & 27.9& 44.5&41.80  \\
    \multirow{2}{*}{ } & P-MapNet (S) & {52.9} & {29.7}  &{46.8}  & {43.13 \color{red}{(+1.33)}}\\
    \multirow{2}{*}{ } & P-MapNet (S+H) & \textbf{53.5} & \textbf{30.1}  &\textbf{47.3}  & \textbf{43.63 \color{red}{(+1.83)}}\\
    \midrule
    \multirow{3}{*}{$120\times 60\,m$} & HDMapNet  & 48.3 & 27.8& 40.0 & 38.70 \\
    \multirow{2}{*}{ } & P-MapNet (S)  &{52.5} &{34.3}  &{48.7}  & {45.17 \color{red}{(+6.47)}}  \\
    \multirow{2}{*}{ } & P-MapNet (S+H)  &\textbf{53.1} &\textbf{34.7}  &\textbf{49.0}  & \textbf{45.60 \color{red}{(+6.90)}}  \\
    \bottomrule
  \end{tabular}
  }
      
  \end{center}
  \label{tab:av2_miou}
  \vspace{-5mm}
\end{table}

\begin{table*}[t]
\vspace{+2mm}
  \caption{\textbf{Perceptual Metric of HDmap Prior.} We utilize the LPIPS metric to evaluate the \textit{realism} of S+H models on $120m\times 60m$ perception range. And the improvements in the S+H setting are more significant compared to those in the S-only setting. }
  \begin{center}
  \resizebox{0.80\textwidth}{!}{
    \begin{tabular}{c|c|ccc|ccc}
    \toprule
    Range & Method &Modality& mIoU$\uparrow$ & LPIPS$\downarrow$ &Modality& mIoU$\uparrow$ & LPIPS$\downarrow$\\
    \midrule
     \multirow{3}{*}{$120\times 60\,m$}&Baseline&C &33.77&  0.8050& C+L &49.07&  0.7872 \\
     &P-MapNet (S)&C &40.33 \color{red}{(+6.56)}&  0.7926 \color{red}{(1.54\%)}&C+L& 60.20 \color{red}{(+11.13)} & 0.7607 \color{red}{(3.37\%)} \\
     &P-MapNet (S+H)&C & \textbf{40.87 
 \color{red}{(+7.10)}}&  \textbf{0.7717 \color{red}{(4.14\%)}}&C+L& \textbf{61.77 \color{red}{(+12.70)}} & \textbf{0.7124 \color{red}{(9.50\%)}} \\
      \midrule
     \multirow{3}{*}{$240\times 60\,m$}&Baseline&C &26.77&  0.8484&C+L &36.47&  0.8408 \\
     &P-MapNet (S)&C& 42.20 \color{red}{(+15.43)} & 0.8192 \color{red}{(3.44\%)}&C+L& 48.87 \color{red}{(+12.40)} & 0.8097 \color{red}{(3.70\%)} \\
     &P-MapNet (S+H)&C& \textbf{45.50 \color{red}{(+18.73)}} & \textbf{0.7906 \color{red}{(6.81\%)}}&C+L& \textbf{49.93 \color{red}{(+13.46)}} & \textbf{0.7765 \color{red}{(7.65\%)}} \\
    \bottomrule
  \end{tabular}
  }
  \end{center}
  \label{tab:lpips}
  \vspace{-6mm}
\end{table*}

\textbf{Perceptual Metric of HDmap Prior.}
The HDMap Priors Module endeavors to map the network output onto the distribution of HDMaps to make it more \textit{realistic}. To evaluate the \textit{realism} of the HDMap prior refinement Module output, we utilize a perceptual metric LPIPS\cite{Lpips} (lower values indicate better performance). The enhancements achieved in the S+H setting are considerably greater  when compared to those in the S-only setting as demonstrated in Tab.\ref{tab:lpips}.

\textbf{Vectorization Results.}
We also conducted a comparison of vectorization results by employing post-processing to obtain vectorized HD Maps. As detailed in Tab.~\ref{tab:mIOU}, we achieve the best instance detection AP results across distance ranges. 

\textbf{Does SDMap pior work for direct vectorzied map prediction?}
As demonstrated in Tab.\ref{tab:maptr}, to confirm the universality of our SDMap prior, we integrated our SDMap Prior Module into MapTR\cite{liao2022maptr} (with only minor modifications), an end-to-end framework, referred to as the MapTR-SDMap method. Our MapTR-SDMap method also led to a significant improvement in mean Average Precision (mAP).
\begin{table}[h]
  \caption{
  \textbf{Comparisons with MapTR\cite{liao2022maptr} on nuScenes \textit{val} set.} We conducted a comparison between MapTR fused with the SDMap prior method (MapTR-SDMap) and the vanilla MapTR\cite{liao2022maptr}, using only surround-view cameras as input and we use predefined CD thresholds of 0.5m, 1.0m and 1.5m. Our results demonstrated superior performance, highlighting the effectiveness of our SDMap prior fusion method.}
  \begin{center}
  \resizebox{0.48\textwidth}{!}{
    \begin{tabular}{c|c|ccc|c}
    \toprule
    Range & Method  & Div. & Ped. & Bound. & mAP\\
    \midrule
    \multirow{3}{*}{$60\times 30\,m$} & MapTR\cite{liao2022maptr} & 49.50 & 41.17 & 51.08 & 47.25  \\
    \multirow{2}{*}{ } & MapTR-SDMap & \textbf{50.92} & \textbf{43.71}  &\textbf{53.49}  & \textbf{49.37}{ (+2.21)}\\
    \multirow{2}{*}{ } & P-MapNet & 26.08 & 17.66 & 48.43 & 30.72 \\
    \midrule
    \multirow{3}{*}{$120\times 60\,m$} & MapTR\cite{liao2022maptr}  & 26.00 & 18.89 & 15.73 & 20.20 \\
    \multirow{2}{*}{ } & MapTR-SDMap  & \textbf{27.23} & {21.95}  & {19.50}  & {22.89 {(+2.69)}}  \\
    \multirow{2}{*}{ } & P-MapNet & {19.50} & \textbf{24.72} & \textbf{42.48} & \textbf{28.90} \\
    \midrule
    \multirow{3}{*}{$240\times 60\,m$} & MapTR\cite{liao2022maptr}  & 12.69 & 7.17 & 4.23 & 8.03 \\
    \multirow{2}{*}{ } & MapTR-SDMap  &\textbf{22.74} & {16.34}  & {10.53}  & {16.53 {(+8.50)}}  \\
    \multirow{2}{*}{ } & P-MapNet & 14.51 & \textbf{25.63} & \textbf{28.11} & \textbf{22.75} \\
    \bottomrule
  \end{tabular}
  }
      
  \end{center}
  \label{tab:maptr}
  \vspace{-5mm}
\end{table}

\begin{figure}[h]
    \centering
    % \vspace{-3mm}
    \includegraphics[width=0.48\textwidth]{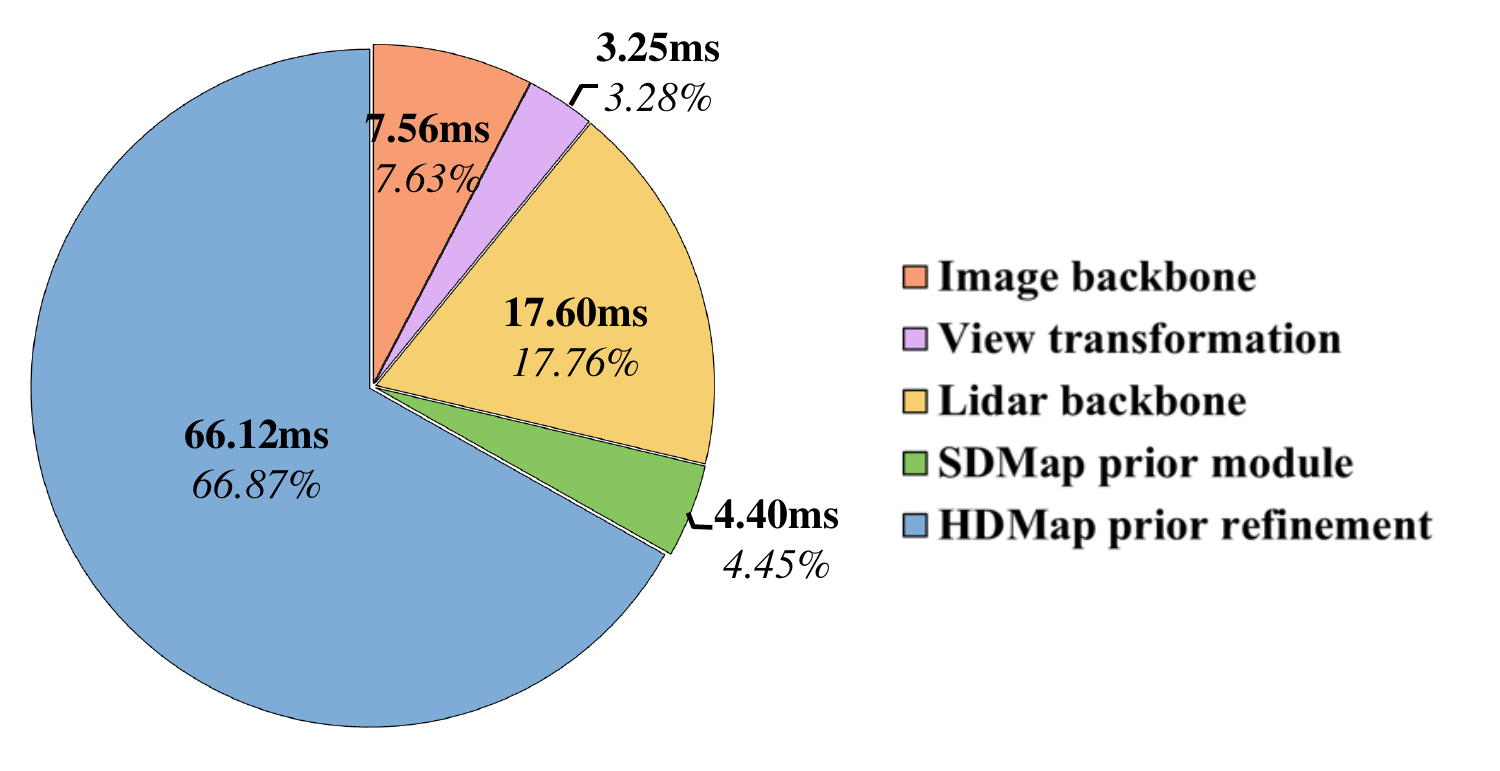}
    \vspace{-8mm}
    \caption{\textbf{Detailed runtime.} 
    We conduct runtime profiling of each component in P-MapNet at a range of 60 × 120m on one RTX 3090 GPU.}
\label{Fig:time}
\vspace{-3mm}
\end{figure}
\subsection{Ablation Study}
All ablative experiments are conducted on nuScenes val set with a perception range of $120m \times 60m$ and the camera-LiDAR fusion(C+L) configuration. 

\textbf{Detailed Runtime.} \label{time}
In Fig.~\ref{Fig:time}, we provide the detailed wallclock runtime of each component in P-MapNet with both camera and LiDAR inputs. As a recap, a complete FPS evaluation is reported in Tab.~\ref{tab:mIOU}. As shown by the profiling, the HDMap prior is computationally heavy but it is indeed optional. Practitioners can switch between the SDmap only setting or the SDMap+HDMap setting, depending on the computation overhead (e.g., onboard or offboard).

\begin{table}[h]
  % \vspace{-2mm}
  \caption{\textbf{Ablations about SD Maps fusion strategies.} The experiments are conducted with range of $120 \times 60m$ and C+L as inputs. ``w/o SDMap" is the baseline \cite{li2022hdmapnet}. ``w/o SDMap, w\ Self.Attn" only employed BEV queries self-attention.}
  \begin{center}

  \resizebox{0.48\textwidth}{!}{
    \begin{tabular}{c|ccc|cccc}
    \toprule
    Fusion Method & Div. & Ped. & Bound. & mIoU\\
    \midrule
    w/o SDMap & 53.2 & 36.9 & 57.1 & 49.07 \\ 
    w/o SDMap, w/ Self.Attn. & 57.7 & 42.0 & 60.6 & 53.43 \\
    Simply-concat & 59.4 & 43.2 & 61.6 &54.73\\ %& 62.8 & 46.4  & 64.9 & 57.37  
    CNN-concat & 60.2 & 45.5  & 63.1 & 56.27 \\ %0.636 0.487 0.664
    Cross.Attn. & \textbf{63.6}  & \textbf{50.2}  & \textbf{66.8} & \textbf{60.20}  \\
    \bottomrule
  \end{tabular}
  }
      
  \end{center}
  \label{tab:fusion_mode}
  \vspace{-6mm}
\end{table}

\textbf{SDMap Prior Fusion Strategies.}
To validate the effectiveness of our proposed fusion approach for SDMap priors, we experimented with various fusion strategies, the details of which are summarized in Tab.~\ref{tab:fusion_mode}. 
In an initial evaluation, a straightforward concatenation (termed "Simple-concat") of the rasterized SDMap with BEV features led to a mIoU boost of about 5\%. 
A better approach, where we exploit CNNs to encode and concatenate the rasterized SDMap, furthered this improvement to about 7\%. 
Nonetheless, the straightforward concatenation techniques were hampered by spatial misalignment issues, preventing the full capitalization of the SDMap priors' potential. Interestingly, leveraging self-attention solely for BEV queries also enhanced performance. Among all the approaches tested, our method anchored on cross-attention demonstrated the most substantial gains.

\textbf{Ablation of BEV-SDPrior Cross-attention Layers.} \label{attention_layers}
As the number of transformer layers increases, performance of our method improves, but it eventually reaches a point of saturation since the SDMap priors contain low-dimensional information, and excessively large network layers are susceptible to overfitting, as shown in Tab.~\ref{tab:attn_layer}.

\begin{table}[h]
  % \vspace{-3mm}
  \caption{\textbf{Ablations about the number of BEV-SDPrior Cross-attention Layers.} During training, we evaluated memory usage with a batch size of 4, while for inference, we measured frames per second (FPS) with a batch size of 1.} 
  \begin{center}
  \vspace{-3mm}\resizebox{0.48\textwidth}{!}{
    \begin{tabular}{c|ccc|ccc}
    \toprule
    Attention Layer& Div. & Ped. & Bound. & mIoU & Memory (GB) & FPS\\
    \midrule
     1 & 62.6 & 48.4 & 65.6 & 58.87 & 19.03  & 19.60  \\%626 0.484 0.656
    2 & \textbf{63.6} & \textbf{50.2} & \textbf{66.8} & \textbf{60.20} & 20.20 & 18.56 \\
    4 & 60.6 & 44.9 & 63.2 & 56.23  & 23.24 &  18.45 \\ %0.606 0.449 0.632
    6 & 58.7 & 42.4 & 61.8 & 54.30  & OOM & - \\ %0.587 0.424 0.618
    \bottomrule
  \end{tabular}
  }
  \end{center}
  \label{tab:attn_layer}
  \vspace{-3mm}
\end{table}

\textbf{The generalization capabitliy of HDMap MAE.} \label{crossdata}
In order to verify the generalizability of our HDMap Prior refinement module, we pre-train on Argoverse2 and nuScenes datasets respectively, and fine-tune on nuScenes dataset and test the prediction results mIOU. The results are shown in the Tab.~\ref{tab:cross-data}, and it can be seen that the model pre-trained on Argoverse2 is only $0.64\%$ mIOU lower than the pre-trained model on nuScenes, which can prove that our refinement module indeed captures the HDMap priors information with generalization capability rather than overfitting to the dataset.
\begin{table}[h]
\vspace{-6mm}
  \caption{
  \textbf{Cross-dataset experiment of HDMap Priors.} We pre-trained the HDMap Prior Module on Argoverse2 and nuScenes datasets, respectively, and tested it on nuScenes val, using a range of $120\times 60m$ and RGB+LidAR inputs.}
  \begin{center}
  \resizebox{0.48\textwidth}{!}{
    \begin{tabular}{c|ccc|c|c}
    \toprule
    Pre-Train Dataset & Div. & Ped. & Bound. & mIoU $\uparrow$ & LPIPS $\downarrow$\\
    \midrule

     Argoverse v2& 64.5 & 51.3 & 67.6 & 61.13 \color{red}{(+0.93)} & 0.7203 \color{red}{(8.49\%)}\\
     Nuscense& 65.3 & 52.0 & 68.0 & 61.77 \color{red}{(+1.57)} & 0.7124 \color{red}{(9.50\%)}\\
    \bottomrule
  \end{tabular}
  }
  \end{center}
  \label{tab:cross-data}
  \vspace{-8mm}
\end{table}

% \input{tables/mask_ratio}
% \textbf{Mask proportion Experiment.} \label{mask_proportion}

% As show in Tab.~\ref{tab:mask_ratio}, we test the effect of using different mask ratios for pre-training on the refinement results, too high a mask ratio will lead to the lack of valid information and the actual refinement process of the input difference is large, too low a mask ratio can not force the network to capture the HDMap priors, we choose the optimal $50\%$ as the ratio of pre-training of our method.

% \textbf{Pretraining and Masking Strategy.}
% We conducted a comparison between results with and without pre-training, which clearly demonstrates that pretraining is effective in capturing HDMap priors.
% We devised two distinct mask strategies: the grid-based strategy and the random-mask strategy. Our approach, which utilized the random sampling strategy, produced the most promising results. 
% For additional details, please refer to Appendix~\ref{appendix:mask}.

\bibliographystyle{plain}  
\bibliography{root} 

\clearpage \appendix 
\section{Appendix Section}\label{append}

\subsection{Implementation Details}
P-MapNet is trained with four NVIDIA GeForce RTX 3090 GPUs. We use the Adam optimizer and StepLR schedule for trainning with a learning rate of $5 \times 10^{-4}$.
For fairness comparison, we adopt the EfficientNet-B0  pretrained on ImageNet as the perspective view image encoder and use a MLP to convert to the BEV features. 
To encode the point clouds for the LiDAR BEV features, we utilize the PointPillars framework, operating at an output dimension of 128.
During the pretraining phase for the HDMap prior, we trained for 20 epochs for each range. Subsequently, we combined the \textit{BEV Feature Fusion} with the \textit{HDMap Prior Refinement} module and conducted an additional 10 epochs of training to obtain the final HDMap predictions. 

\subsection{Further Study on SDmap Prior}
\subsubsection{Integrating SDMap Prior into end-to-end vectorized framework}\label{appendix:vec_sd}

As demonstrated in Tab.\ref{tab:maptr}, to confirm the universality of our SDMap prior, we integrated our SDMap Prior Module into MapTR (with only minor modifications), a state-of-the-art end-to-end framework, referred to as the MapTR-SDMap method. Our MapTR-SDMap method also led to a significant improvement in mean Average Precision (mAP).
%Our initial investigations confirmed the performance of the end-to-end framework in long-range scenarios. 

\begin{figure}[h]
    \centering
    \includegraphics[width=1.0\linewidth]{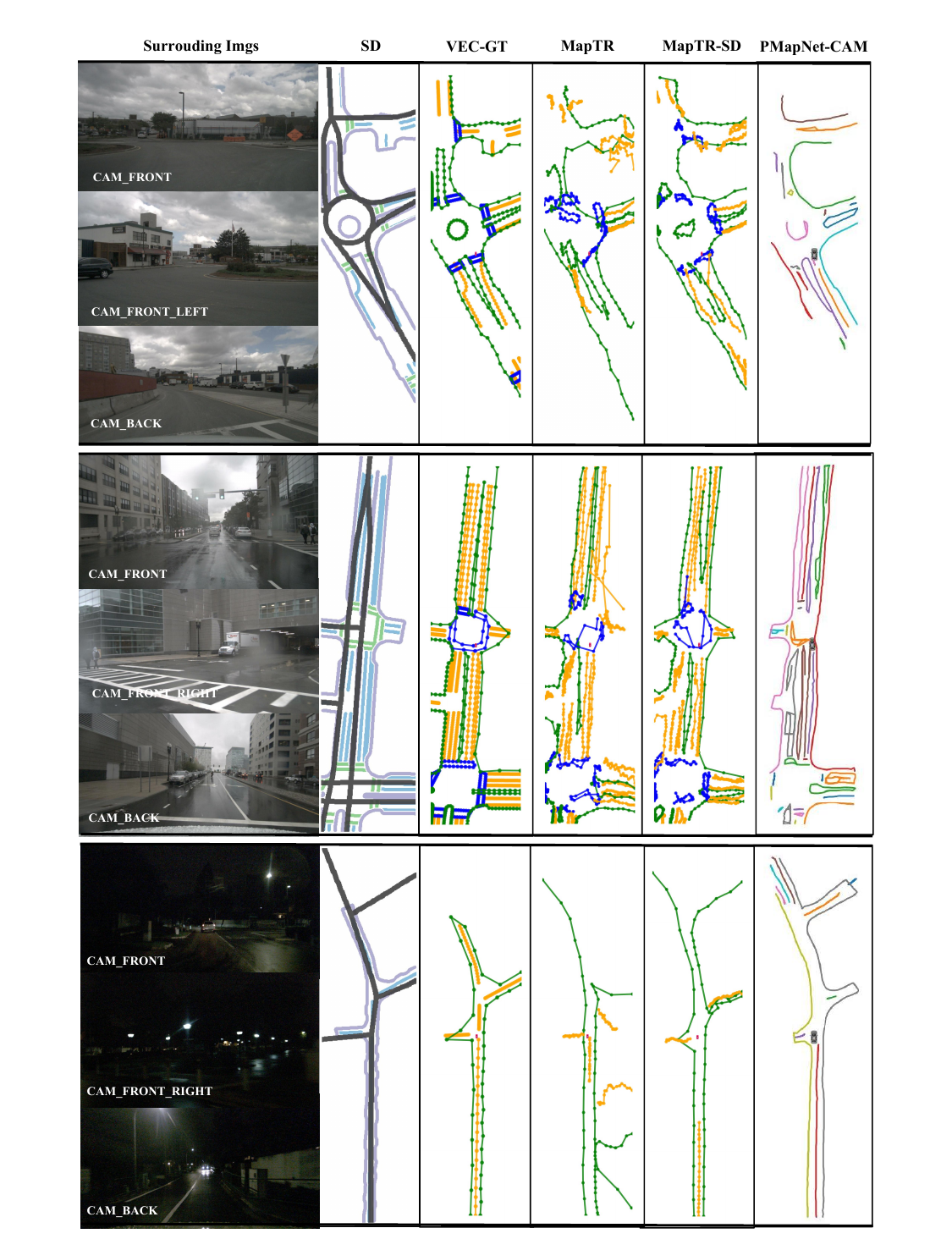}
    \caption{\textbf{The qualitative results of vectorized results on a perception range of $ 240m \times 60m$ .} We integrate the SDMap Prior Module into the MapTR (with only minor modifications), referred to as the MapTR-SD. PMapNet-CAM is our method with both SDMap prior and HDMap prior utilizing the post process.}
    \label{vector}
\end{figure}

The visualization results in Fig.~\ref{vector} also show that MapTR-SDMap performs better under the most challenging $240\times 60m$ ultra-long range of perception. It can also be seen that the segmentation-post-processing approach has stable results because it is dense-prediction, while the end-to-end vectorization approach still has some challenges such as significant predictive bias and challenges in keypoint selection. In conclusion, our proposed SDMap Prior fusion method demonstrates performance improvement in both the segmentation-postprocessing framework and the end-to-end framework.

\begin{table}[h]
  \caption{
  \textbf{Comparisons with MapTR on nuScenes \textit{val} set.} We conducted a comparison between MapTR fused with the SDMap prior method (MapTR-SDMap) and the vanilla MapTR, using only surround-view cameras as input and we use predefined CD thresholds of 0.5m, 1.0m and 1.5m. Our results demonstrated superior performance, highlighting the effectiveness of our SDMap prior fusion method.}
  \begin{center}
  \resizebox{0.48\textwidth}{!}{
    \begin{tabular}{c|c|ccc|c}
    \toprule
    Range & Method  & Div. & Ped. & Bound. & mAP\\
    \midrule
    \multirow{3}{*}{$60\times 30\,m$} & MapTR & 49.50 & 41.17 & 51.08 & 47.25  \\
    \multirow{2}{*}{ } & MapTR-SDMap & \textbf{50.92} & \textbf{43.71}  &\textbf{53.49}  & \textbf{49.37}{ (+2.21)}\\
    \multirow{2}{*}{ } & P-MapNet & 26.08 & 17.66 & 48.43 & 30.72 \\
    \midrule
    \multirow{3}{*}{$120\times 60\,m$} & MapTR  & 26.00 & 18.89 & 15.73 & 20.20 \\
    \multirow{2}{*}{ } & MapTR-SDMap  & \textbf{27.23} & {21.95}  & {19.50}  & {22.89 {(+2.69)}}  \\
    \multirow{2}{*}{ } & P-MapNet & {19.50} & \textbf{24.72} & \textbf{42.48} & \textbf{28.90} \\
    \midrule
    \multirow{3}{*}{$240\times 60\,m$} & MapTR  & 12.69 & 7.17 & 4.23 & 8.03 \\
    \multirow{2}{*}{ } & MapTR-SDMap  &\textbf{22.74} & {16.34}  & {10.53}  & {16.53 {(+8.50)}}  \\
    \multirow{2}{*}{ } & P-MapNet & 14.51 & \textbf{25.63} & \textbf{28.11} & \textbf{22.75} \\
    \bottomrule
  \end{tabular}
  }
      
  \end{center}
  \label{tab:maptr}
  \vspace{-5mm}
\end{table}

\subsubsection{Inconsistencies between Ground Truth and SDMaps}\label{appendix:sd_inconsistence}
\textbf{Influence of Inconsistencies between Ground Truth and SDMaps.} Our SDMap priors are derived from OpenStreetMap (OSM). Nonetheless, due to discrepancies between labeled datasets and actual real-world scenarios, not all roads datasets are comprehensively annotated. This leads to incongruities between the SDMap and HDMap.
Upon a closer examination of OSM, we noticed that there is a category in OSM called \emph{service} road, which is for access roads leading to or located within an industrial estate, camp site, business park, car park, alleys, etc.

Incorporating \emph{service} category roads can enrich the SDMap prior information with greater details. However, it also implies a potential increase in inconsistencies with dataset annotations.
In light of this, we take ablation experiments to determine the advisability of incorporating \emph{service} category roads.

\begin{figure}[h]
    \centering
    \includegraphics[width=1.0\linewidth]{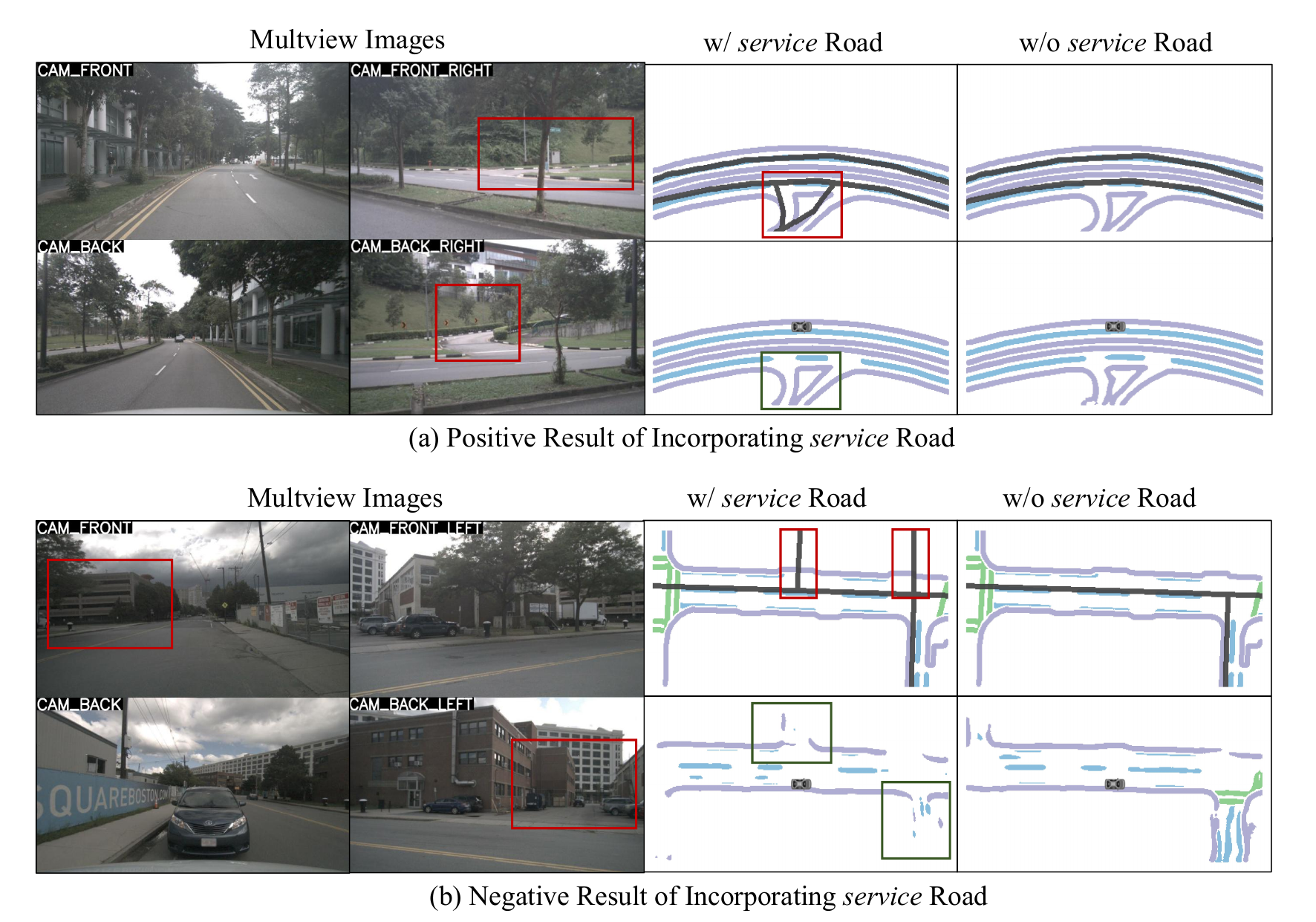}
    \caption{The two scenarios underscore the effects of discrepancies between the ground truth and SDMaps. (b) shows that the network filters most of the \emph{service} roads as noise since majority of them are incongruous with the ground truth, which affects the performance of trunk roads. SDMap prior exhibits commendable efficacy when \emph{service} roads are not introduced. (a) demonstrates that when the distribution of the \emph{service} road deviates from the norm, its performance is enhanced since the network refrains from filtering it out as noise.}
    \label{fig:service}
\end{figure}

As shown in Fig.~\ref{fig:service}, we select two cases to demonstrate the impact of inconsistencies between datasets map annotations and SDMaps. Specifically, in the Fig. \ref{fig:service}(a), the inclusion of a \emph{service} roads (an industrial estate) yields a positive outcome, where the SDMap aligns well with the ground truth dataset.

Nevertheless, in most cases, SDMaps with \emph{service} roads are inconsistent with the ground truth of datasets, primarily due to the lack of detailed annotations for internal roads.
% since they does not annotate very detailed internal roads.
% In that way, The network learns the distribution of most service roads and filters them as noise.
Consequently, during the training process, the network learns the distribution of most service roads and filters them as noise. This inadvertently led to some primary roads being erroneously filtered out as noise.
As depicted in Fig. \ref{fig:service}(b), the service road (two alleys) highlighted in the red box is absent in the ground truth. The network perceives it as noise and consequently does not produce the corresponding road. However, it also neglects to generate a road for the primary route indicated in the ground truth, delineated within the green box, resulting in a significant discrepancy.
Conversely, the network that excludes \emph{service} roads avoids learning numerous erroneous SDMap distributions. This enables the network to more effectively assimilate SDMap information pertaining to main roads, even though many detailed SDMaps may be missing.
The visualization in the right side of Fig. \ref{fig:service}(b) demonstrates that the SDMap prior effectively guides the generation of HDMap. It even reconstructs the pedestrian crosswalks and lane at remote intersections, even though these reconstructions do not align with the actual ground truth.

% osm with service road 
\begin{table}[h]
  \vspace{-4mm}
  \caption{\textbf{Quantitative results on different OSM category.} Incorporating \emph{service} road introduce richer information but also involve inconsistency. In terms of segmentation mIoU results,
   the absence of \emph{service} roads in the SDMap prior leads to an improvement of approximately 2\% in performance.}
  \begin{center}
  \resizebox{0.48\textwidth}{!}{
    \begin{tabular}{c|ccc|cccc}
    \toprule
    With \emph{service} road & Divider & Ped Crossing & Boundary & mIoU\\
    \midrule
    w/ \emph{service} & 62.4 & 47.9 & 65.3 & 58.53 \\
    w/o \emph{service} & \textbf{63.6} & \textbf{50.2} & \textbf{66.8} & \textbf{60.20} \\
    \bottomrule
  \end{tabular}
  }
      
  \end{center}
  \label{tab:service}
  \vspace{-4mm}
\end{table}
In terms of quantitative metrics, as show in Tab. \ref{tab:service}, excluding \emph{service} roads results in a 2\% mIoU improvement. The modest difference in these metrics suggests that the network can effectively filter out noise when introduced to numerous SDMaps that deviate from the ground truth. It further emphasizes the effectiveness of an SDMap focused on main roads in guiding the generation of the HDMap.
% , the quantitative result is shown in Tab. \ref{tab:service}.

\begin{figure}[h]
    \centering
    \includegraphics[width=1.0\linewidth]{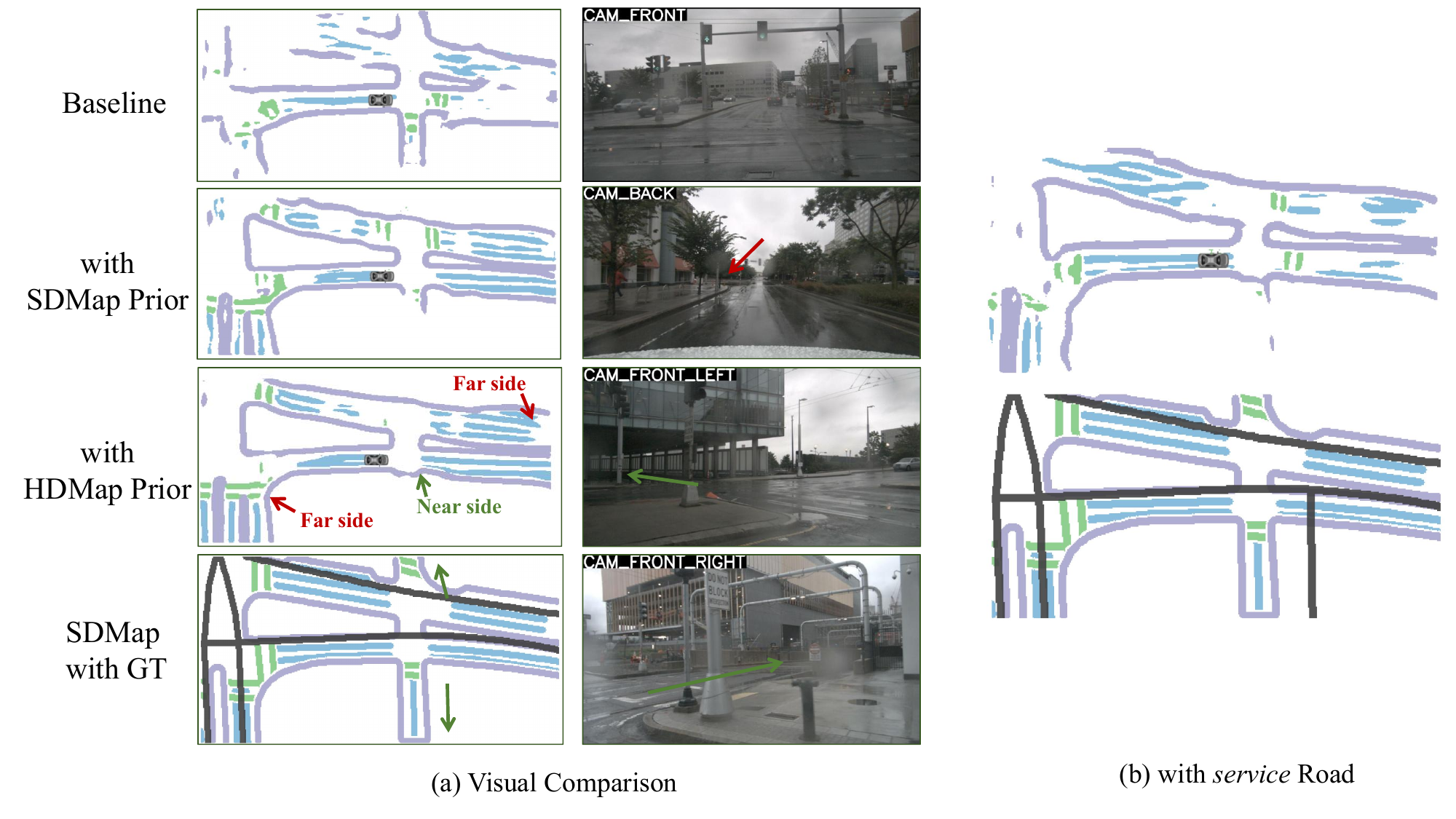}
    \caption{\textbf{The negative results in the near side fork roads.} (a) demonstrates that the baseline exhibits proficient performance in predicting near side forks. Yet, due to the SDMap's exclusive focus on main roads, the prediction accuracy for near side forks diminishes upon integrating both the SDMap and HDMap prior information. (b) shows that even if \emph{service} road information is added, the network will filter out this SDMap as noise.}
    \label{fig:osm_vis}
\end{figure}

\textbf{Visualization Analysis of the Inconsistencies between Ground Truth and SDMaps.}
As seen in Fig. \ref{fig:osm_vis}, we select a case to show the negative result in the near side due to the inconsistencies. 
Obviously, the baseline shows that when SDMap prior information is not integrated, both the left and right forks at the near side can be predicted, but the far side cannot be predicted clearly due to weather reasons and visual distance.

When leveraging the SDMap prior to bolster HDMap generation, the predictions for the near side forks roads deteriorate due to the SDMap's exclusive focus on trunk roads. Furthermore, incorporating the HDMap prior alleviates artifacts and bridges the gaps, but this inadvertently diminishes prediction performance on the near side fork roads, as shown in Fig. \ref{fig:osm_vis}(a).

However, we also validate the case using the model with \emph{service} roads as in Fig. \ref{fig:osm_vis}(b). The network perceives the \emph{service} SDMap of the fork road (to the industrial park) as noise and filters it out, as mentioned in the previous part. 
The other left side fork road is not contained in the \emph{service}.
Consequently, it performs suboptimal as well since the SDMaps is not detailed enough and the discrepancies between SDMaps and dataset's ground truth.

In summary, we introduce SDMap priors information and undertaken a comprehensive analysis of several intriguing cases. Our intent is to offer insights that may inspire future researchers to further the utilization of SDMap priors in HDMap generation.

\subsubsection{Ablation of BEV Feature Downsampling Factor} \label{appendix:feature_map_size}
Different downsampling factor $d$ impact the size of feature map $\mathcal{B}_{small}$ in the fusion module. Larger feature maps convey more information but can result in higher GPU memory usage and slower inference speed. As shown in Tab.~\ref{tab:feature_map_size}, to strike a balance between speed and accuracy, we opt for a size of $50 \times 25$.

\begin{table}[h]
  \caption{
  \textbf{Ablations about downsampling factor.} We conducted a comparison of mIOU results for feature sizes at a range of $120\times 60m$ with different down-sampling multiples. The term ``OOM" represents the GPU memory is exhausted. During training, we evaluated memory usage with a batch size of 4, while for inference, we measured frames per second (FPS) with a batch size of 1.
  }
  
  \begin{center}
  \resizebox{0.48\textwidth}{!}{
    \begin{tabular}{cc|ccc|ccc}
    \toprule
    Factor & Feature Map Size & Div. & Ped. & Bound. & mIoU & Memory(GB) & FPS\\
    \midrule
    $d=2$ & $100\times 50$ &-  &-  &-  &- & OOM & -  \\
    $d=4$ & $50\times 25$ & \textbf{63.6} & \textbf{50.2} & \textbf{66.8} & \textbf{60.20}  & 20.2 & 18.56 \\
    $d=8$ & $25\times 12$ & 60.7 & 45.0 & 63.3 & 56.33 & 18.3 &  19.60 \\
    \bottomrule
  \end{tabular}
  }
      
  \end{center}
  \label{tab:feature_map_size}
  \vspace{-5mm}
\end{table}

\begin{table}[h]
  \caption{\textbf{Ablations about mask proportion.} We use different random mask ratios for pre-training, with higher mask ratios being harder for the reconstruction.
  }
  \begin{center}
  \resizebox{0.48\textwidth}{!}{
    \begin{tabular}{c|ccc|c}
    \toprule
    Mask Proportion & Divider & Ped Crossing & Boundary & mIoU\\
    \midrule
    25\% & 64.8 & 51.4 & 67.6 & 61.27 \\
    50\% & \textbf{65.3} & {52.0} & \textbf{68.0} & \textbf{61.77} \\
    75\% & {64.7} & \textbf{52.1} & {67.7} & {61.50} \\
    \bottomrule
  \end{tabular}
  }
  \end{center}
  \label{tab:mask_ratio}
  \vspace{-4mm}
\end{table}

\subsection{Further Study on HDmap Prior}
\label{appendix:hdmap}

\subsubsection{Mask proportion Experiment} \label{appendix:mask_proportion}

As show in Tab.~\ref{tab:mask_ratio}, we test the effect of using different mask ratios for pre-training on the refinement results, too high a mask ratio will lead to the lack of valid information and the actual refinement process of the input difference is large, too low a mask ratio can not force the network to capture the HDMap priors, we choose the optimal $50\%$ as the ratio of pre-training of our method.

\vspace{-3mm}
\begin{figure}[h]
    \centering
    \includegraphics[width=0.75\linewidth]{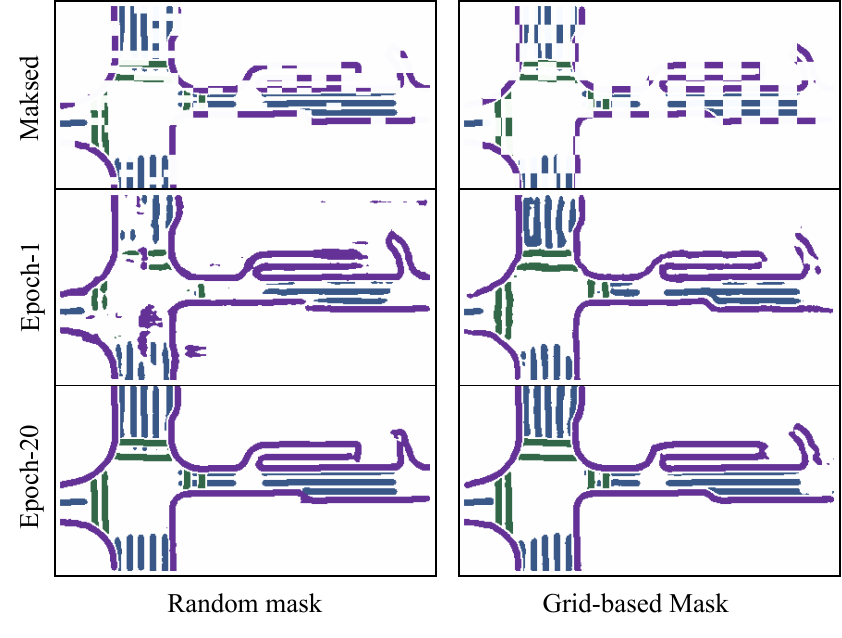}
    \caption{\textbf{Different mask strategies. }``Maksd" refers to the pre-training inputs after applying various masking strategies, and ``Epoch-1" and ``Epoch-20" denote the reconstruction results at the first and twentieth epochs of the pre-training process, respectively.}
    \label{fig:mask}
\end{figure}
\subsubsection{Ablation of Mask Strategy} \label{appendix:mask}
The grid-based strategy uses a patch size of $20 \times 20$ pixels and
keeps one of every two patches. And the random-mask strategy selects one of the patch sizes from $20\times20$, $20\times40$, $25\times50$, or $40\times80$ with a 50\% probability for masking. The visualization results are presented in Figure~\ref{fig:mask}. With pre-training, the refinement module effectively learns the HDMap priors. As depicted in Tab.\ref{tab:mask}, our approach employing the random sampling strategy yielded the most favorable results.

\begin{table}[h]
  \caption{\textbf{Ablations about mask strategy.} ``w/o pretrain" signifies that we do not pre-train the HDMap Prior Refinement module. Interestingly, our random-mask method yields superior results in this context.}
  \begin{center}
  \resizebox{0.48\textwidth}{!}{
    \begin{tabular}{c|ccc|c}
    \toprule
    Mask Strategy & Divider & Ped Crossing & Boundary & mIoU\\
    \midrule
    w/o Pre-train & 64.1 & 51.4 & 67.4 & 60.97 \\
    Gird-based & 65.1 & \textbf{52.3} & 67.8 & 61.73 \\
    Random-mask & \textbf{65.3} & {52.0} & \textbf{68.0} & \textbf{61.77} \\
    \bottomrule
  \end{tabular}
  }
  \end{center}
  \label{tab:mask}
  % \vspace{-4mm}
\end{table}

\begin{figure}[!h]
\vspace{-10mm}
    \centering
    \includegraphics[width=0.77\linewidth]{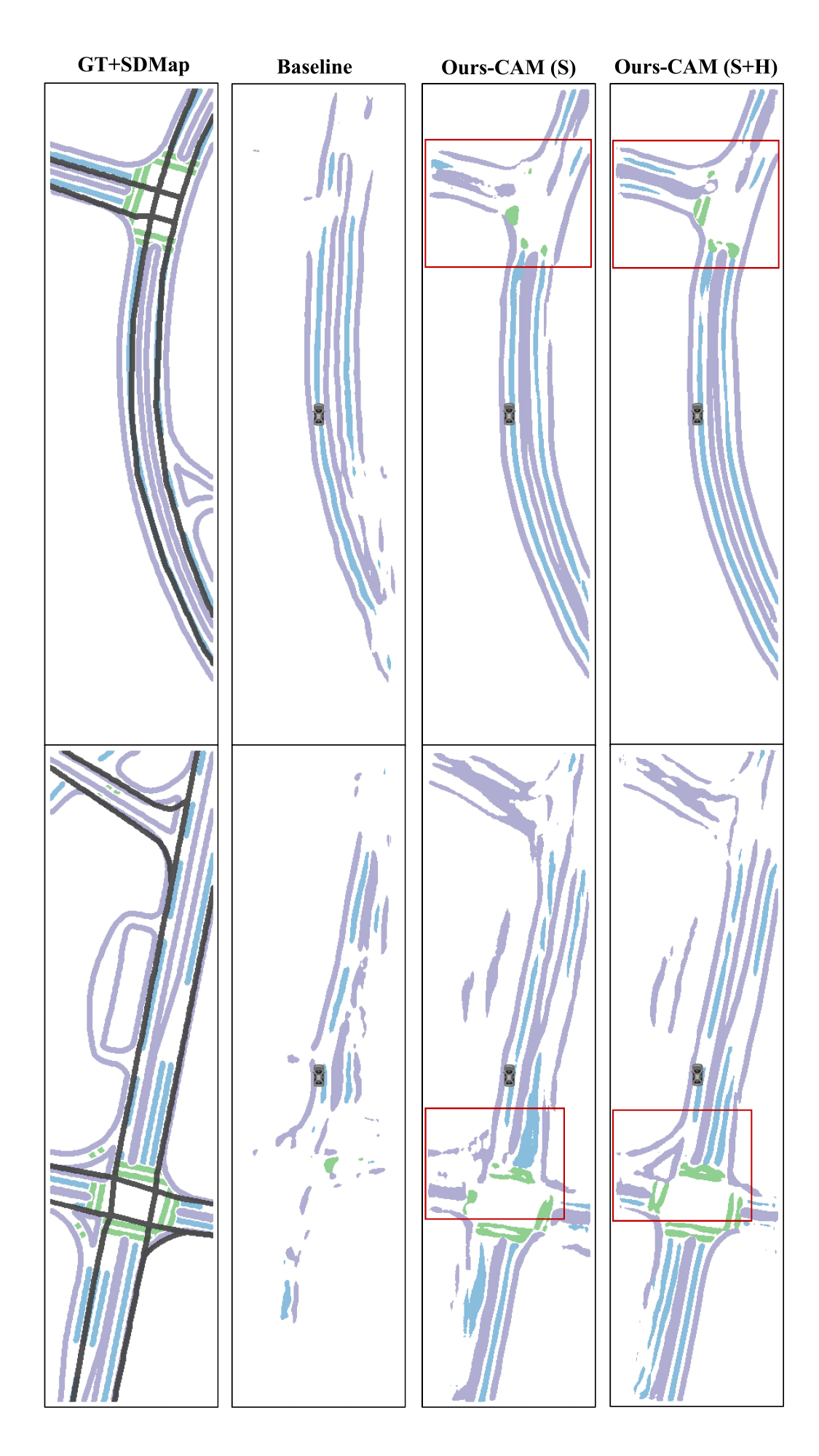}
    \caption{\textbf{The qualitative results of only camera method on a perception range of $240m \times 60m$.} The SDMap Prior Module improves the prediction results by fusing road structure priors. While HDMap Prior Module makes it closer to the distribution of HDMap to a certain extent, making it look more realistic.  }
    \label{fig:cam mae}
\end{figure}

\subsection{QUALITATIVE VISUALIZATION} \label{appendix_vis}

\subsubsection{Segmentation Qualitative Results}

As depicted in the Fig.\ref{fig:cam mae} and Fig.~\ref{vector}, we provide additional perceptual results under diverse weather conditions, and our method exhibits superior performance.

\subsubsection{SD Map Data Visualization} \label{appendix:sd_map}
\begin{figure}[h]
  \vspace{-10mm}
\centering
\includegraphics[width=0.50\textwidth]{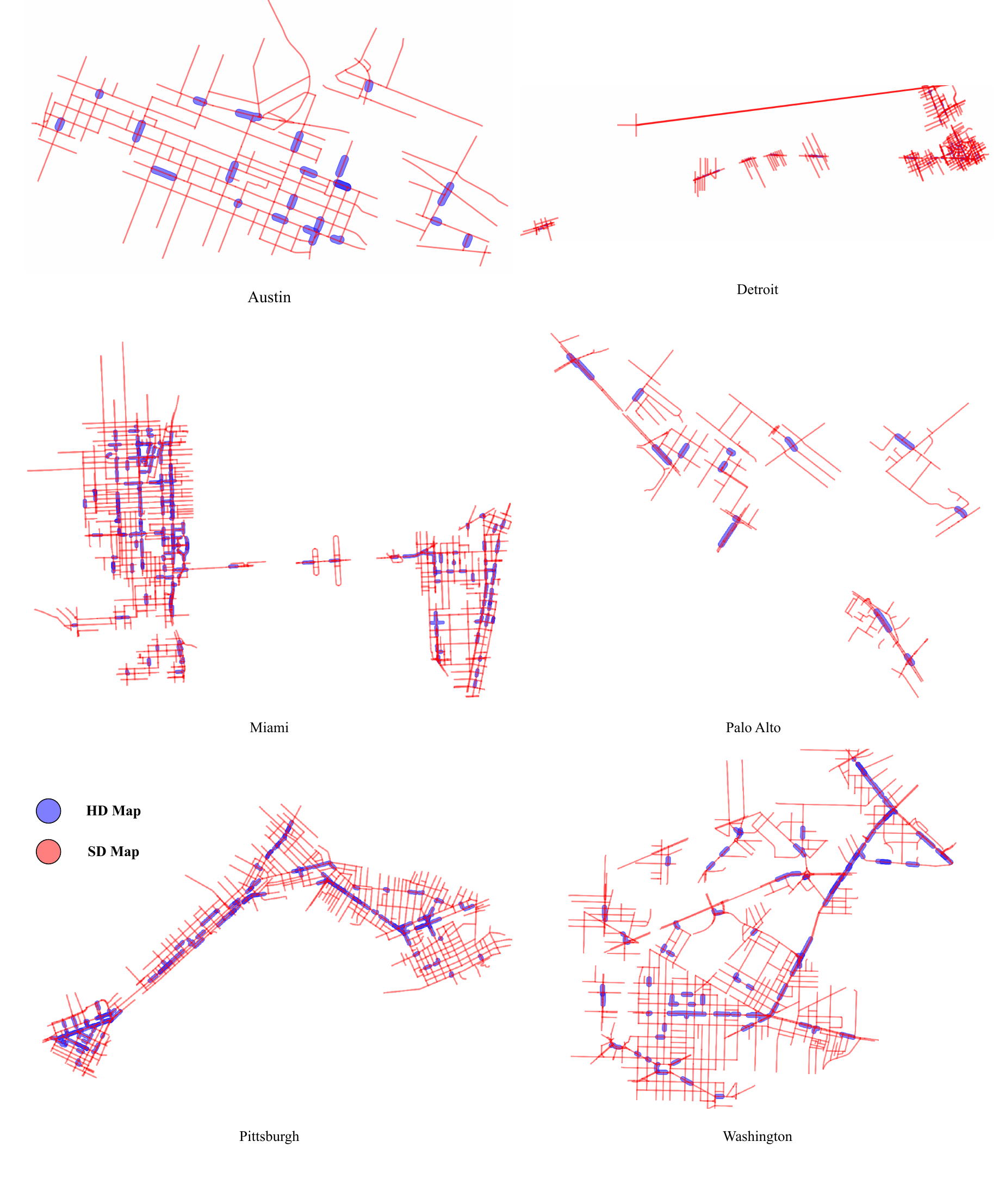}
\caption{The visualizations of SD Map data and HD Map data on the Argoverse2 dataset.}
\label{fig:appendix_argo2}
\end{figure}

\begin{figure}[t]
  \vspace{-5mm}
\centering
\includegraphics[width=0.50\textwidth]{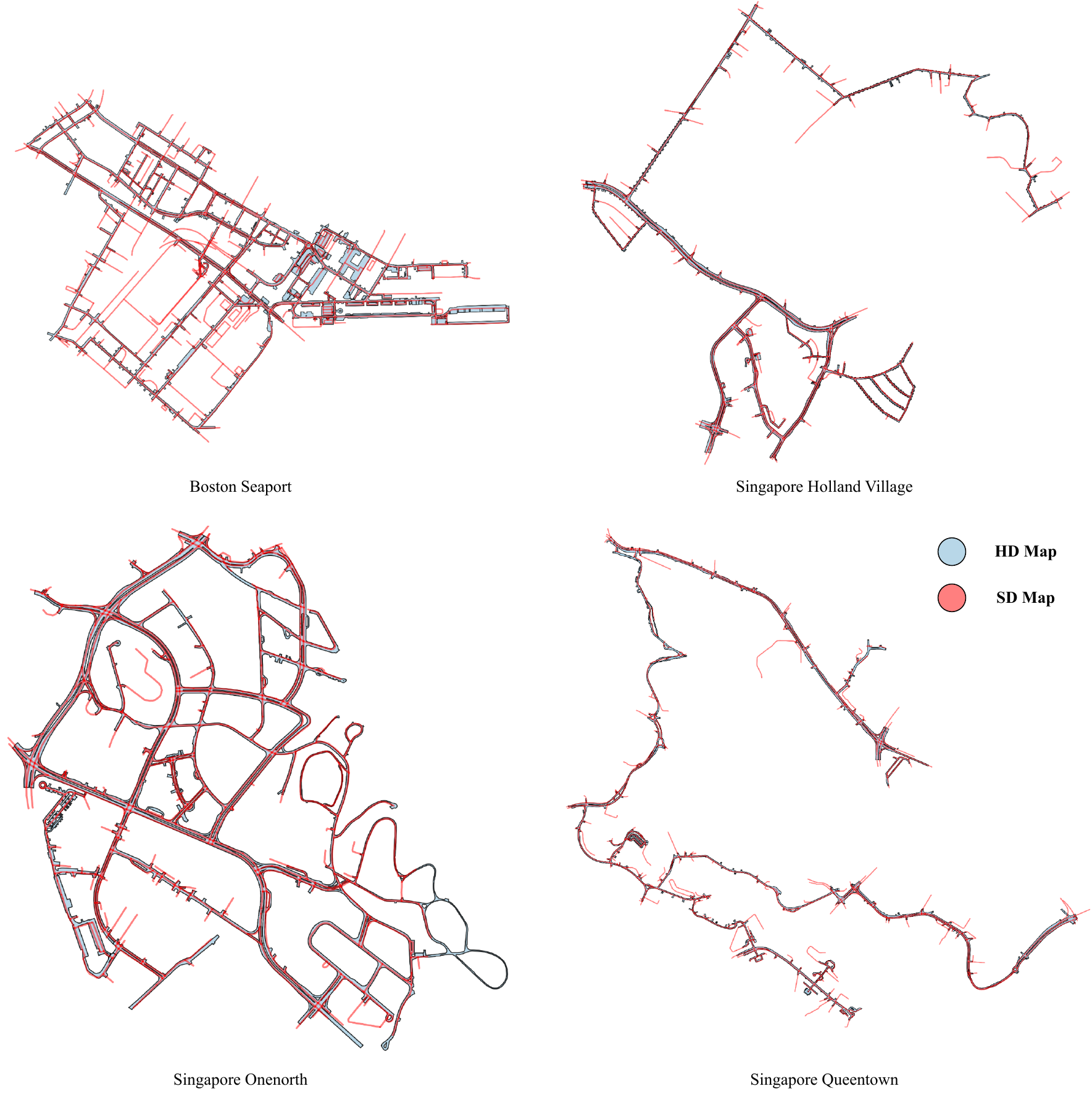}
% \vspace{-2mm}
\caption{The visualizations of SD Map data and HD Map data on the nuScenes dataset.}
\label{fig:appendix_nuscenes}
% \vspace{-12mm}
\end{figure}

We supplemented the SD Map data on both the  and  datasets, the specific details are outlined in Tab.~\ref{tab:sd_details}. The visualization of SD map data and HD Map data facilitated by the dataset, is presented in Fig.~\ref{fig:appendix_argo2} and Fig.~\ref{fig:appendix_nuscenes}.

\begin{table}[h]
  % \vspace{-12mm}
  \caption{\textbf{SDMap data details.} In order to generate the SDMap data, we extract the \textit{road}, \textit{road link} and \textit{special road} data from the \textit{highway section} of OSM data and perform coordinate alignment and data filtering.}
  \begin{center}
  \resizebox{0.48\textwidth}{!}{
    \begin{tabular}{c|c|ccc}
    \toprule
    Dataset & Sub-Map & Lane Numbers & Total Length(km)\\
    \midrule
    \multirow{4}{*}{NuScenes} &  Singapore-OneNorth & 576 & 23.4   \\
    \multirow{4}{*}{ } & Singapore-HollandVillage & 359 & 16.9  \\ 
    \multirow{4}{*}{ } & Singapore-Queenstown     & 393 & 17.9  \\
    \multirow{4}{*}{ } & Boston-Seaport            & 342 & 32.1  \\
    \midrule
    \multirow{6}{*}{Argoverse2} & Austin & 193 & 46.5   \\
    \multirow{6}{*}{ } & Detroit & 853 & 160.6  \\ 
    \multirow{6}{*}{ } & Miami  & 1226 & 178.2  \\
    \multirow{6}{*}{ } & Palo Alto & 315 & 33.4  \\
    \multirow{6}{*}{ } & Pittsburgh & 640 & 112.3  \\
    \multirow{6}{*}{ } & Washington DC & 1020 & 150.6  \\
    \bottomrule
  \end{tabular}
  }
  \end{center}
  \label{tab:sd_details}
  \vspace{-8mm}
\end{table}

\vspace{12pt}

\end{document}